\documentclass[10pt,twocolumn,letterpaper]{article}

\usepackage[pagenumbers]{cvpr} 
\usepackage{times}
\usepackage{epsfig}
\usepackage{graphicx}
\usepackage{amsmath}
\usepackage{amssymb}
\usepackage{enumitem}
\usepackage{comment}
\usepackage{color}
\usepackage{caption}
\usepackage{booktabs}
\usepackage{siunitx} %
\usepackage{multirow}
\usepackage{gensymb}
\usepackage{ulem}
\usepackage{xspace}
\usepackage{arydshln}
\usepackage[accsupp]{axessibility} %
\usepackage[dvipsnames]{xcolor}

\def\eg{\emph{e.g.,}\xspace}
\def\ie{\emph{i.e.,}\xspace}
\def\etal{\emph{et al.}\xspace}
\definecolor{cvprblue}{rgb}{0.21,0.49,0.74}
\usepackage[pagebackref,breaklinks,colorlinks,citecolor=cvprblue]{hyperref}

\title{EventEgo3D: 3D Human Motion Capture from Egocentric Event Streams} 

\author{\hspace{-25pt}Christen Millerdurai\textsuperscript{\text{1,2}} 
\qquad\quad
Hiroyasu Akada\textsuperscript{1} 
\qquad\qquad
Jian Wang\textsuperscript{1} \\
\hspace{-10pt}\qquad
Diogo Luvizon\textsuperscript{1} 
\qquad\quad\hspace{5pt}
Christian Theobalt\textsuperscript{1} 
\qquad
Vladislav Golyanik\textsuperscript{1} \\
\textsuperscript{1}MPI for Informatics, SIC 
\qquad
\textsuperscript{2}Saarland University, SIC 
}

\begin{document}

\twocolumn[{%
\maketitle

\renewcommand\twocolumn[1][]{#1}%
\begin{center}
    \captionsetup{type=figure}
    \includegraphics[width=\linewidth]{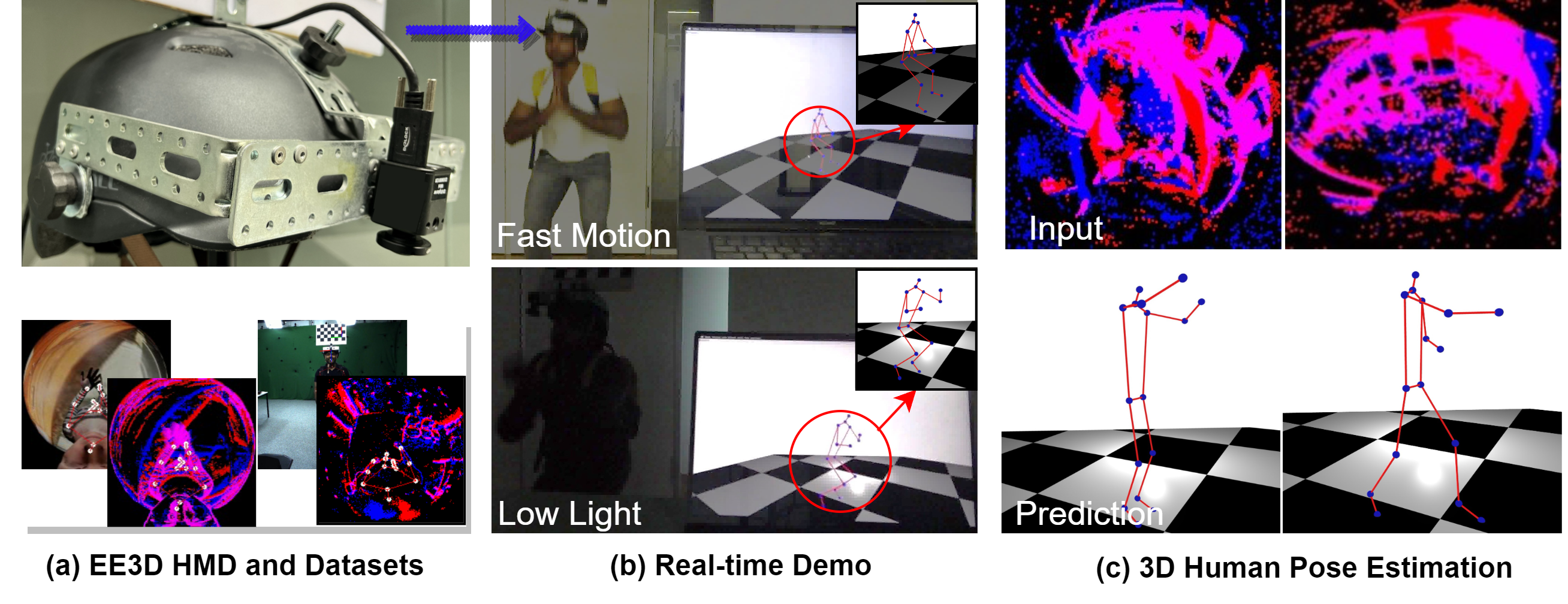}
    \vspace{-10pt} 
   \caption{
   \textbf{EventEgo3D is the first approach for real-time 3D human motion capture from egocentric event streams:}  (a) A photograph of our new head-mounted device (HDM) with a custom-designed egocentric fisheye event camera (top) and visualisations of our synthetically rendered dataset and a real dataset recorded with the HDM (bottom); (b) Real-time demo achieving the pose update rate of $140$Hz; (c) Visualisation of real event streams (top) and the corresponding 3D human poses from a third-person perspective. 
   } 
\label{fig:teaser}
\end{center}%
}]
\begin{abstract} 
    Monocular egocentric 3D human motion capture is a challenging and actively researched problem. 
    Existing methods use synchronously operating visual sensors (e.g.~RGB cameras) and 
    often fail under low lighting and fast motions, which can be restricting in many applications involving head-mounted devices. 
    In response to the existing limitations, 
    this paper 1) introduces a new problem, i.e., 3D human motion capture from an egocentric monocular event camera with a fisheye lens, and 2) proposes the first approach to it called EventEgo3D (EE3D). 
    Event streams have high temporal resolution and provide reliable cues for 3D human motion capture under 
    high-speed human motions and rapidly changing illumination. 
    The proposed EE3D framework is specifically tailored for learning with event streams in 
    the LNES 
    representation, 
    enabling high 3D reconstruction accuracy. 
    We also design a prototype of a mobile head-mounted device with an event camera and record a real dataset with event observations and the ground-truth 3D human poses (in addition to the synthetic dataset). 
    Our EE3D demonstrates robustness and superior 3D accuracy compared to existing solutions across various challenging experiments while supporting 
    real-time 3D pose update rates of $140$Hz.\footnote{\url{https://4dqv.mpi-inf.mpg.de/EventEgo3D/}} 
\end{abstract}

\vspace{-15pt}
\section{Introduction}\label{sec:intro}

Head-mounted devices (HMD) have a high potential to become the next mobile and pervasive computing platform in human society that could enable many applications in education, driving or personal assistance systems, gaming, and many others. 
HMDs enable increased flexibility 
and allow users to move freely and explore the environments they live and work in. 
Consequently, egocentric 3D human pose estimation 
became an active research field 
during the last few years, 
with several works 
focusing on 
recovering 
3D human poses from down-facing fisheye RGB cameras installed on an HMD \cite{rhodin2016egocap, xu2019mo2cap2, zhao2021egoglass, wang2022estimating, hakada2022unrealego, wang2023scene, wang2021estimating, Tom2023SelfPose3E, Liu2023, Li2023EgoBody}. 

Existing egocentric setups have been predominantly demonstrated in the literature with monocular RGB cameras with a fisheye lens. 
While these experimental prototypes showed high 3D human pose estimation accuracy under certain assumptions, 
monocular RGB cameras on HMDs have multiple fundamental disadvantages: They are prone to over- or under-exposure and motion blur in the presence of high-speed human motions; consume comparably much power for a mobile device;  
moreover, they record the image frames synchronously and require constantly high data processing throughput. 
Hence, our work is motivated by the observation that 
multiple disadvantages of RGB-based HMDs 
can be alleviated with a different type of visual sensor, \textit{i.e.}, event cameras. 
Event cameras record streams of events, \textit{i.e.}, asynchronous per-pixel brightness 
changes at high (${\mu}$s) temporal resolution. 
They also support an increased dynamic range and consume less power (on the order of tens of $m$W) than average RGB cameras (consuming Watts) \cite{gallego2020event}. 
Also, no events are triggered if there are no changes in the scene (apart from noisy signals). 

Note that existing RGB-based (especially learning-based) techniques cannot be re-purposed for event streams in a straightforward way; new and dedicated approaches are required to unveil all the 
advantages of event cameras. 
We are thus inspired by the recent 
progress in event-based 3D reconstruction 
in different scenarios 
\cite{EventCap2020, rudnev2021eventhands, Zou2021eventhpe, jiang2023evhandpose}. 
However, an egocentric HMD setup utilising an event camera with a fisheye lens has not been previously addressed in the literature. 
This configuration introduces additional challenges, including the need for lightweight design to accommodate high-speed human motions (real-time processing) and the significant amount of background events generated by the \textit{moving event camera}.

This paper addresses the challenges associated with the design of such an HMD with a monocular egocentric event camera with a fisheye lens; see Fig.~\ref{fig:teaser} for an overview. 
We introduce a prototypical design of a compact HMD that can be worn by a human and used under fast motions, with the processing happening on a laptop carried in a backpack (Fig.~\ref{fig:teaser}-(a) on top). 
We then propose a lightweight neural network architecture operating on a suitable event stream representation (LNES \cite{rudnev2021eventhands}) for real-time 
performance. 
Our method, EventEgo3D or EE3D for short, encodes the incoming events in the compact representation and decodes 2D heatmaps of the observed human joint locations. 
Afterwards, the lifting block regresses the 3D human poses. 
We also propose a residual event propagation module specifically designed for the egocentric monocular setting that highlights the wearer's human amongst the background events and 
provides reliable predictions even under the lack of events due to the absence of human motion.

Due to the lack of event datasets in the egocentric setting, we build a large-scale synthetic dataset for training our approach. 
Moreover, we record a real dataset with 3D ground-truth human poses and 2D event stream observations with our real HMD (Fig.~\ref{fig:teaser}-(a)).  
The proposed real-world dataset allows for fine-tuning methods trained on the synthetic dataset and boosting the accuracy of the egocentric event-based 3D pose estimation in real-world scenarios.

In summary, this paper defines a new problem, \textit{i.e.}, 3D human pose estimation from a monocular egocentric event camera, and makes 
the following technical contributions: 
\begin{itemize} 
    \item EE3D, the first end-to-end trainable neural approach for 3D human motion capture from an egocentric event camera with a fisheye lens installed on a mobile head-mounted device (Sec.~\ref{sec:method});
    \item Lightweight Residual Event Propagation Module and Egocentric Pose Module tailored for real-time 3D human motion capture (pose update rate of $140$Hz) from egocentric event streams with high 3D reconstruction accuracy;
    \item The design of a compact head-mounted device with an egocentric event camera along with real and synthetic datasets for method training and evaluation (Sec.~\ref{sec:setup_datasets}). 
\end{itemize} 

Our experiments demonstrate 
higher 3D reconstruction accuracy of EE3D in challenging scenarios with high-speed human motions from egocentric monocular event streams 
compared to 
existing (closely related) methods. 
The significance of each proposed module is evaluated and confirmed in an ablative study (Sec.~\ref{sec:experiments}). 

\section{Related Work}
\label{sec:related_work} 
We next review related methods for egocentric 3D human pose estimation and event-based 3D reconstruction. 

\subsection{Egocentric 3D Human Pose Estimation}
3D human pose estimation from egocentric monocular or stereo RGB views has been actively studied during the last decade. 
While the earliest approaches were optimisation-based \cite{rhodin2016egocap}, the field promptly adopted neural architectures following the state of the art in human pose estimation. 
Thus, follow-up methods used a two-stream CNN architecture \cite{xu2019mo2cap2} and auto-encoders for monocular \cite{tome2019xr, Tom2023SelfPose3E} and stereo inputs \cite{zhao2021egoglass, hakada2022unrealego,hakada2024unrealego2}. 
Another work focused on the automatic calibration of fisheye cameras widely used in the egocentric setting \cite{zhang2021automatic}. 
Recent papers leverage human motion priors and temporal constraints for predictions in the global coordinate frame \cite{wang2021estimating}; reinforcement learning for improved physical plausibility of the estimated motions \cite{yuan2019ego, luo2021dynamics}; semi-supervised GAN-based human pose enhancement with external views \cite{wang2022estimating} and depth estimation \cite{wang2023scene}; and scene-conditioned denoising diffusion probabilistic models~\cite{Zhang_2023_ICCV}. 
Khirodkar et al.~\cite{Khirodkar_2023_ICCV} address a slightly different setting and use a multi-stream transformer to capture multiple humans in front-facing egocentric views. 

All these works demonstrated promising results and pushed the field forward. 
They, however, were designed for synchronously operating RGB cameras and, hence---as every RGB-based method---suffer from inherent limitations of these sensors (detailed in Sec.~\ref{sec:intro}). 
Thus, only a few of them support real-time frame rates \cite{xu2019mo2cap2, tome2019xr}. 
Moreover, it is unreasonable to expect that RGB-based approaches can be easily adapted for event streams. 
In contrast, we propose an approach 
that (for the first time) accounts for the new data type in the context of egocentric 3D vision (events) and estimates 3D human poses at high 3D pose update rates. 

Last but not least, none of the existing datasets for the training and evaluation of egocentric 3D human pose estimation techniques and related problems \cite{rhodin2016egocap, xu2019mo2cap2, tome2019xr, wang2021estimating, Zhang_ECCV_2022, wang2023scene, Pan_2023_ICCV, Khirodkar_2023_ICCV} provide event streams or frames at framerate sufficient to generate events with event steam simulators \cite{rebecq18a_esim}. 
To evaluate and train our EE3D approach, we synthesise and record the necessary datasets (\textit{i.e.}, synthetic, real and background augmentation) required to investigate event-based 3D human pose estimation on HMDs.

\subsection{Event-based Methods for 3D Reconstruction} 
Substantial discrepancy between RGB frames and asynchronous events 
has recently led to the emergence of dedicated event-based or hybrid 3D techniques for humans \cite{EventCap2020, Zou2021eventhpe, chen2022efficient}, hands \cite{rudnev2021eventhands, xue2022event} and general objects \cite{Nehvi2021, Wang2022EvAC3D, rudnev2023eventnerf}.

Nehvi et al.~\cite{Nehvi2021} track non-rigid 3D objects (polygonal meshes or parametric 3D models) with 
a differentiable event stream simulator. 
Rudnev et al.~\cite{rudnev2021eventhands} synthesise a 
dataset with human hands 
to train a neural 3D hand pose tracker with a Kalman filter. 
They introduce a lightweight LNES representation of events for learning as an improvement upon event frames. 
Next, Xue et al.~\cite{xue2022event} optimise the parameters of a 3D hand model by associating events with mesh faces using the expectation-maximisation framework assuming that events are predominantly triggered by hand contours. 
Some works \cite{EventCap2020, Zou2021eventhpe} use RGB frames along with the event streams, and some represent events as spatiotemporal points in space and encode them either as point clouds 
\cite{chen2022efficient, Millerdurai_3DV2024}. 
Consequently, most of these approaches are slow (due to different reasons such as iterative optimisation or computationally expensive operations on 3D point clouds), with the notable exception of EventHands \cite{rudnev2021eventhands} 
achieving up to $1$kHz hand pose update rates.

We use LNES \cite{rudnev2021eventhands} as it is independent of the input event count,  
facilitates real-time inference and can be efficiently processed with 
neural components 
(e.g.~CNN layers). 
In contrast to all approaches discussed above, our method is specifically designed for the egocentric setting and
achieves the highest accuracy level among all compared methods. 
This is achieved by incorporating 
a novel residual %
mechanism that propagates events (event history) from the previous frame to the current one, prioritising events triggered around the human. 
This is also helpful 
when only a few events are triggered due to the lack of motion.

\begin{figure*}[t]
\begin{center}
   \includegraphics[width=1.0\textwidth]{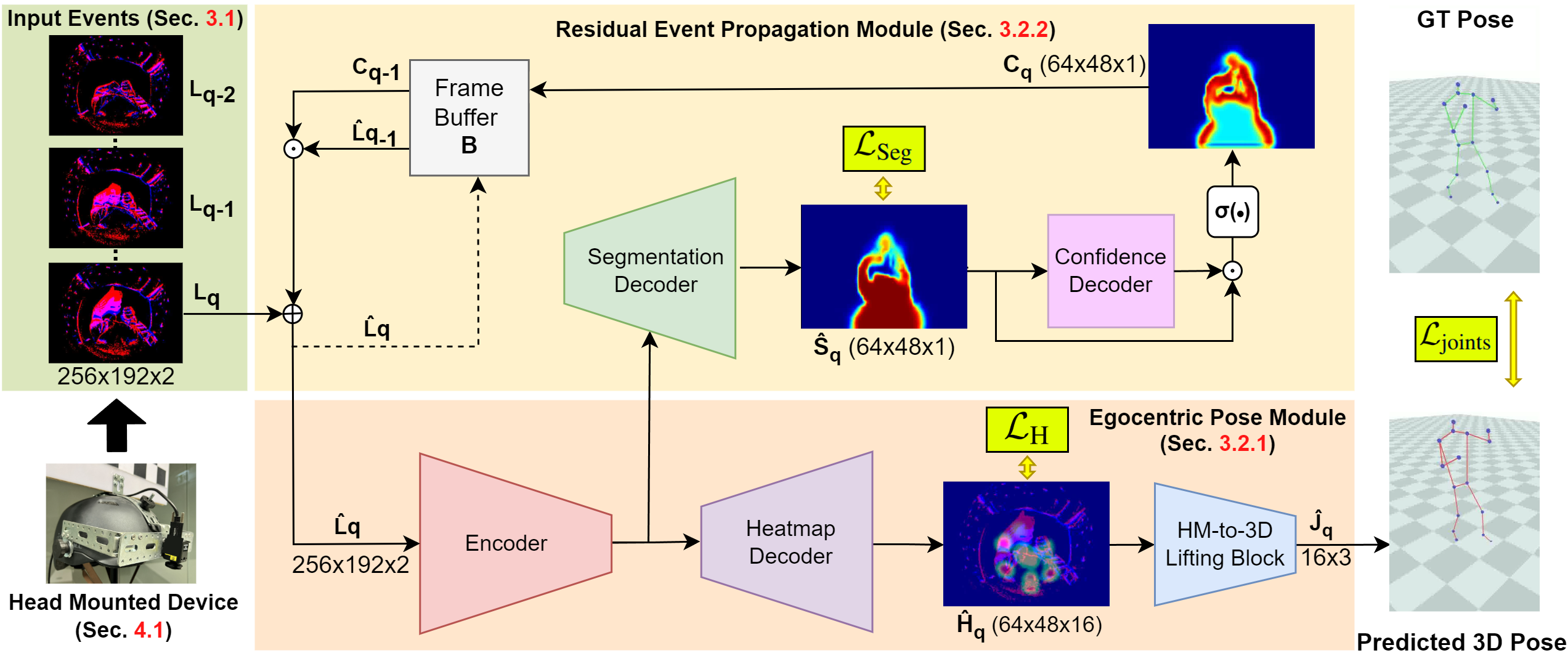}
\end{center}
    \vspace{-10pt} 
    \caption{
    \textbf{Overview of our EventEgo3D approach}. The HMD captures an egocentric event stream converted to a series of 2D LNES frames \cite{rudnev2021eventhands}, 
    from which our neural architecture 
    regresses the 3D poses of the HMD user. 
    The residual event propagation module (REPM) emphasises events triggered around the human by considering the temporal context of observations (realised with a frame buffer with event decay based on event confidence). 
    REPM, hence, helps the encoder-decoder (from LNES to heatmaps) and the heatmap lifting module to estimate accurate 3D human poses. 
    The method is supervised with ground-truth segmentations, heatmaps and 3D human poses. 
}
\label{fig:eventego}
\end{figure*}

\section{The EventEgo3D Approach}\label{sec:method} 
Our EE3D approach estimates 3D human poses from an egocentric monocular event camera with a fisheye lens.
We first explain the event camera model in Sec.~\ref{subsec:camera_model} and then describe the proposed framework in Sec.~\ref{subsec:EgoEvents3D}.

\subsection{Event Camera Preliminaries}
\label{subsec:camera_model}
Event cameras %
capture event streams, \ie a 1D temporal sequence that contains discrete packets of asynchronous events that indicate the brightness change of a pixel of the sensor. 
In our case, we use a fisheye lens and the Scaramuzza projection model~\cite{scaramuzza2006toolbox} for it, introducing a wider field of view required for our HMD. 
An event is a tuple of the form $e_i = (x_i, y_i, t_i, p_i)$ with the $i$-th index representing the event fired at 
pixel location $(x_i, y_i)$ with its corresponding timestamp $t_i$ and a polarity $p_i \in \{-1, 1\}$. 
The timestamps $t_i$ of modern event cameras have $\mu$s temporal resolution. 
The event 
is generated when the change in logarithmic brightness $\mathbb{L}$ at the pixel location ($x_i, y_i$) exceeds a predefined threshold $C$, 
\textit{i.e.}, $|\mathbb{L}(x_i,y_i,t_i)-\mathbb{L}(x_i,y_i,t_i-t_{p})|\geq{C}$, where $t_p$ represents the
previous triggering time at the same pixel location.
$p = -1$ indicates that the brightness has decreased by $C$; otherwise, it has increased if $p = 1$.

Modern neural 3D computer vision architectures~\cite{rudnev2021eventhands, lan2023tracking, jiang2023evhandpose} require event streams to be converted to a regular representation, usually 2D or 3D. 
To this end, we adopt the locally normalised event surfaces (LNES) \cite{rudnev2021eventhands} that aggregate the 
event tuples into a compact 2D representation as a function of time windows. 
A time window of size $T$ is constructed by collecting all events between the first event $e_0$ (relative to the given time window) and $e_k$, where $t_k - t_0 \leq T$. 
The events from the time window are stored in the 2D LNES frame $\mathbf{L} \in \mathbb{R}^{H \times W \times 2}$.
The LNES frame is updated by $L(x_i, y_i, p_i) = \frac{t_i - t_0}{T}$, with $i \in \{1,\hdots,k\}$, and where each event triggered at pixel location ($x$, $y$) updates the corresponding pixel location of the LNES frame. 

\subsection{Architecture of EventEgo3D} 
\label{subsec:EgoEvents3D}
Our approach takes $N$ consecutive LNES frames ${\mathbf{B}=\{\mathbf{L}_1, \hdots, \mathbf{L}_N\}}$, $\mathbf{L}_q{\in} \mathbb{R}^{192 \times 256 \times 2}$ %
as input and regresses the 3D human body pose per each LNES denoted by 
${\mathbf{O}=\{\mathbf{\hat{J}}_1, \hdots, \mathbf{\hat{J}}_N\}}$,  $\mathbf{\hat{J}}_q{\in}\mathbb{R}^{16 \times 3}$; $q \in \{1, \hdots, N\}$.
$\mathbf{\hat{J}}_q$ include the joints of the head, neck, shoulders, elbows, wrists, hips, knees, ankles, and feet. 
\par
The proposed framework includes two modules; see Fig.~\ref{fig:eventego}. 
First, the Egocentric Pose Module (EPM)
estimates the 3D coordinates of human body joints.
Subsequently, the Residual Event Propagation Module (REPM) propagates events from the previous LNES frame to the current one. 
The REPM module allows the framework 1) to focus more on the events triggered around the human (than those of the background) and 2) to retain the 3D human pose when only a few events are generated due to the absence of motions.

\subsubsection{Egocentric Pose Module (EPM)}
We regress 3D joints from the input 
$\mathbf{L}_q$ 
in two steps: 1) 2D joint heatmap estimation and 2) the heatmap-to-3D lifting. 
\par
\noindent \textbf{2D Joint Heatmap Estimation}. 
To estimate the 2D joint heatmaps, we develop a U-Net-based architecture~\cite{ronneberger2015u}. 
Here, we utilise the Blaze blocks~\cite{bazarevsky2020blazepose} as layers of the encoder and decoder to achieve real-time performance. 
The encoder consists of six layers, and the heatmap decoder has four layers. 
Please see Appendix~\ref{sec:impl_details} for more details. 
We use 
$\mathbf{L}_q$ 
to estimate 2D heatmaps of 16 body joints ${\mathbf{\hat H}_q} \in \mathbb{R}^{48 \times 64 \times 16}$. %
The final estimated heatmaps are derived by averaging the intermediate heatmaps. 
The network at this stage is supervised using the mean square error (MSE) between the ground-truth heatmaps and the predicted ones: 
\begin{equation}
\label{eq:losshms}
\mathcal{L}_{\text{H}} = \frac{1}{M_J}\sum_{j=1}^{M_J} \lVert \mathbf{\hat H}_{q,j} - \mathbf{H}_{q,j} \rVert ^2, 
\end{equation}
where $\mathbf{\hat H}_{q,j}$ and $\mathbf{H}_{q,j}$ are the predicted and ground-truth heatmaps of the $j$-th joint; $M_J$ is the number of body joints. 
\par
\noindent \textbf{Heatmap-to-3D Lifting Module}.
Following previous works \cite{tome2019xr, pavlakos2018learning}, the Heatmap-to-3D (HM-to-3D) lifting module takes the estimated heatmaps as input and outputs the 3D joints $\mathbf{\hat J_q} \in \mathbb{R}^{16 \times 3}$. 
The HM-to-3D is a six-layer network containing convolutional blocks and two dense blocks.
The HM-to-3D lifting module is supervised using %
MSE 
between the ground-truth device-centric joint coordinates and the predicted ones at the frame $q$: 
\begin{equation}
\label{eq:lossj3d}
\mathcal{L}_{\text{joints}} = \frac{1}{M_J}\sum_{j=1}^{M_J} \lVert \mathbf{\hat J}_{q,j} - \mathbf{J}_{q,j} \rVert  ^2, 
\end{equation}
where $M_J$ is the number of body joints, and $\mathbf{\hat J}_{j}$ and $\mathbf{J}_{j}$ are the predicted and ground-truth $j$-th joints, respectively.

\subsubsection{Residual Event Propagation Module (REPM)} 
In contrast to 
stationary camera setups, egocentric cameras mounted on HMDs undergo significant movements. 
In the case of our mobile device, 
motion results in a comparably high number of events generated by the background. 
Hence, our approach has to be robust to the background events.
We introduce the Residual Event Propagation Module (REPM) that allows the network to focus on the events generated by the subject wearing the HMD as well as rely on the information from previous frames when few events are observed, \ie when the movement of the human body is small.
REPM comprises the segmentation decoder, the confidence decoder, and the frame buffer \textbf{B}. 
The segmentation decoder first estimates the human body mask. 
Then, the confidence decoder produces feature maps that act on the human body mask to produce confidence maps that indicate regions of the egocentric view to place more importance on. 
Lastly, the frame buffer \textbf{B} stores the past input frame and its corresponding confidence map, providing weighting to important regions of the current frame (see the top part of Fig.~\ref{fig:eventego}).

\par
\noindent \textbf{Segmentation Decoder}.
The segmentation decoder estimates the human body mask $\mathbf{\hat S}_q \in \mathbb{R}^{48 \times 64 \times 1}$ of the HMD user in the egocentric LNES views.
The architectures of this module and the heatmap decoder are the same except for the final layer that outputs human body masks. 

We use the feature maps from multiple levels of the encoder as the input to the segmentation decoder.
The segmentation decoder is supervised by the cross-entropy loss: 
\begin{equation}
\label{eq:loss_seg}
\mathcal{L}_{\text{seg}} = - \mathbf{S}_q \log(\mathbf{\hat S}_q) + (1 - \mathbf{S}_q)\log(1 - \mathbf{\hat S}_q), 
\end{equation}
where $\mathbf{\hat S}_q$ and $\mathbf{S}_q$ are the predicted and ground-truth segmentation masks, respectively. 
\par

\begin{figure}[t]
\begin{center}
   \includegraphics[width=0.9\columnwidth]{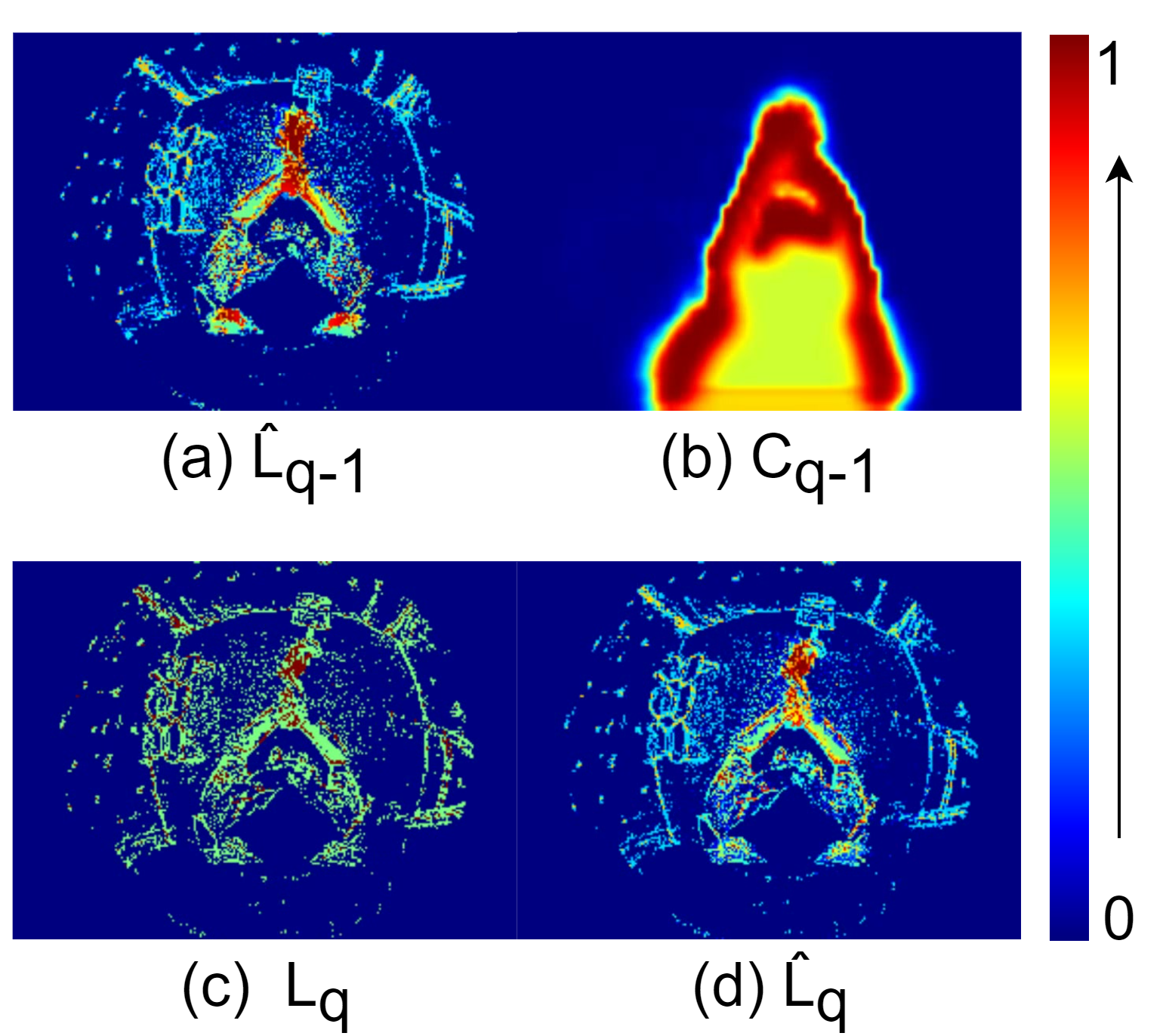}
\end{center}
    \vspace{-10pt} 
    \caption{
        The frame buffer holds previous input frame $\mathbf{\hat{L}}_{q-1}$ (a) and previous confident map $\mathbf{C}_{q-1}$ (b). The $\mathbf{\hat{L}}_{q-1}$ is weighted with $\mathbf{C}_{q-1}$ and added to the current LNES frame $\mathbf{L}_q$ (c) to produce $\mathbf{\hat L}_{q}$ (d). 
        We can observe that the events generated by the subject are highlighted more compared to the background events, thereby prioritising events generated by the subject. 
    } 
\label{fig:ed_l_inp}
\end{figure}
\noindent \textbf{Confidence Decoder}.
The confidence decoder is a four-layer convolution network that takes the human body mask $\mathbf{\hat S}_q$ as input and produces a feature map $\mathbf{S_F}_q \in \mathbb{R}^{48 \times 64 \times 1}$ that, in turn, is used in combination with $\mathbf{\hat S}_q$ to produce the confidence map $\mathbf{C}_q \in \mathbb{R}^{48 \times 64 \times 1}$: 
\begin{equation}
\label{eq:decay_eqn}
\mathbf{C}_q = \operatorname{sigmoid} (\mathbf{\hat S}_q \odot \mathbf{S_F}_q),  
\end{equation}
where ``$\operatorname{sigmoid}(\cdot)$'' is a sigmoid operation and ``$\odot$'' is an element-wise multiplication.

\par
\noindent \textbf{Frame Buffer}. 
The frame buffer $\mathbf{B}$ stores the confidence map $\mathbf{C}_{q-1} \in \mathbb{R}^{48 \times 64 \times 1}$ and input frame $\mathbf{\hat{L}}_{q-1} \in \mathbb{R}^{192 \times 256 \times 2} $ of the previous LNES frame. 
Note that we initialize the frame buffer with zeros at the first frame. 
To compute the current input frame $\mathbf{\hat{L}}_{q}$, we retrieve $\mathbf{C}_{q-1} $ and $\mathbf{\hat{L}}_{q-1}$ using the following expression:  
\begin{equation}
\label{eq:ed_output}
\mathbf{\hat{L}}_{q} = \mathbf{\hat{L}}_{q-1} \odot  \mathbf{C}_{q-1} \oplus \mathbf{L}_q
\end{equation}
where $\mathbf{L}_q$ denotes the LNES frame at the current time, ``$\odot$'' represents an element-wise multiplication, and ``$\oplus$'' represents an element-wise addition.
We normalize the values of $\mathbf{\hat{L}}_{q}$ to the range of $ [-1,1]$. Note, 
$\mathbf{C}_{q-1}$ is resized to $192 \times 256$ before applying Eqn.~\eqref{eq:ed_output}.
See Fig.~\ref{fig:ed_l_inp} for an exemplary visualisation of the components used in Eqn.~\eqref{eq:ed_output}. 
\subsubsection{Loss Terms and Supervision} 
Our method is supervised by the heatmap loss $\mathcal{L_{\text{H}}}$, the segmentation loss $\mathcal{L_{\text{seg}}}$ and the joint loss $\mathcal{L_{\text{joints}}}$. 
Overall, our total loss function is as follows: 
\begin{equation}
\mathcal{L} = 
  \lambda_{\text{joints}} \mathcal{L}_{\text{joints}} 
+  \lambda_{\text{H}} \mathcal{L}_{\text{H}}
+  \lambda_{\text{seg}} \mathcal{L}_{\text{seg}},
\end{equation}
where we set the weight of each loss as $\lambda_{\text{joints}}{=}0.01$, $\lambda_{\text{H}}{=}10$, $\lambda_{\text{seg}}{=}1$.

\section{Our Egocentric Setup and Datasets}\label{sec:setup_datasets}

\begin{figure}[t]
\centering
   \includegraphics[width=0.99\linewidth]{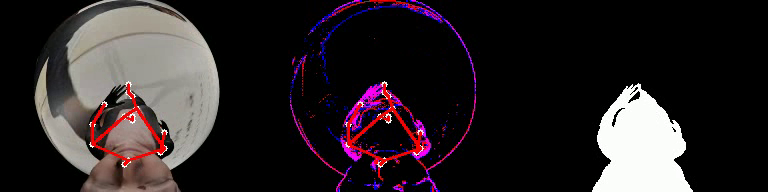}
   \caption{
    Sample from EE3D-S with synthetic RGB image (left), generated event stream (middle), and human body mask (right).
} 
\label{fig:dataset_synt}
\end{figure}
\begin{figure}[t]
\centering
   \includegraphics[width=0.99\linewidth]{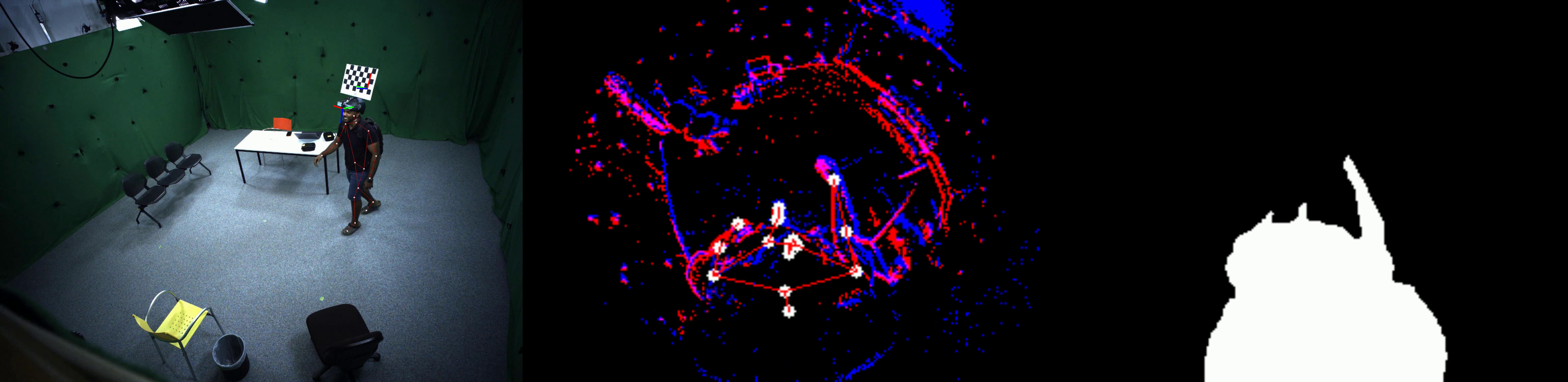}
   \caption{
    Sample from EE3D-R with motion tracking setup (left) used for obtaining the ground-truth 3D poses, event stream (middle), and human body mask (right).
} 
\label{fig:dataset_real}
\end{figure}

We develop a portable head-mounted device (HMD; Fig.~\ref{fig:teaser}-(a)) and capture a real dataset with it. 
\subsection{Head Mounted Device}
Our HMD is a prototypical device consisting of 
a bicycle helmet with a DVXplorer Mini \cite{DVXplorer_Mini} event camera attached to the helmet $3.5$cm away from the user's head; the strap allows a firm attachment on the head. 
We use a fisheye lens Lensagon BF10M14522S118C \cite{Lensation} with a field of view of $190\degree$. 
The total weight of the device is ${\approx}0.42$kg. 
The device is used with a laptop in a backpack for external power supply and real-time on-device computing. 
The compact design and the flexibility of our HMD allow users to freely move their heads and perform rapid human motions. 
\subsection{EE3D Datasets}
We propose two datasets for method training and evaluation: 1) EE3D-S, the large-scale synthetic dataset (used for pre-training), 
and 2) EE3D-R, a real-world dataset capture with our HMD; see 
Figs.~\ref{fig:dataset_synt} and~\ref{fig:dataset_real}. 
Both datasets provide event data, human body masks, and ground-truth 3D poses. 
\subsubsection{EE3D-S (Synthetic)} 
EE3D-S is generated in two steps. 
We use the synthetic egocentric renderings from Xu~\etal~\cite{xu2019mo2cap2} with SMPL-based~\cite{SMPL:2015} virtual humans wearing the virtual copy of our HDM and performing a wide range of motion sequences. 
We render the egocentric views at $480$ frames per second (FPS) and feed them into VID2E~\cite{Gehrig_2020_CVPR}, generating the event streams for each sequence. 
Here, the SMPL~\cite{SMPL:2015} body parameters are linearly interpolated to render views at the desired frame rate. 
We simulate different illuminations by incorporating four point-light sources positioned within a 5-meter radius of the HMD, whose position and light intensity 
randomly change for each sequence. 
In total, we generate $946$ motion sequences with $6.21 \cdot 10^6$ 3D human poses.

\subsubsection{EE3D-R (Real)} 
EE3D-R requires three steps. 
We ask twelve subjects---persons with different body shapes and skin tones---to wear our HMD and perform different motions (e.g.~fast) in 
a multi-view studio 
with $29$ RGB cameras recording at $50$ FPS. 
We capture twelve sequences per subject 
with the following motions: walking, crouching, pushups, boxing, kicking, dancing, interaction with the environment, 
crawling,
sports and jumping. 
We track the 6DoF HMD poses and the human poses in the global reference frame using a multi-view motion capture system \cite{captury}; see Fig.~\ref{fig:dataset_real}. 

Next, we pose the SMPL meshes using the tracked space skeletons and obtain the human body masks by re-projecting the former to the egocentric views. 
Finally, we obtain the tracked 3D human poses in the local coordinate system of the HMD. 
In total, we obtain $4.64 \cdot 10^5$ poses spanning around $155$ minutes of recordings. 

\noindent \textbf{HDM Calibration.} 
To transform the tracked 3D human poses into the world coordinate frame, 
we need to estimate the 6DoF pose of HMD in it. 
We use a common image-based calibration policy as follows. 
To obtain the chequerboard images for the hand-eye calibration procedure, events are generated from the chequerboard and subsequently converted to images using E2VID \cite{Rebecq19pami}. 
We generate events uniformly across the chequerboard by sliding the chequerboard diagonally. 
The final position of the chequerboard after this sliding motion serves as the required chequerboard position for hand-eye calibration. 
The obtained transformation matrix from the hand-eye calibration is used to transform 3D poses and SMPL body meshes to the local coordinate system of the HMD.

\noindent \textbf{Event Augmentation.}
Motions and data recorded in the multi-view studio would not allow satisfactory generalisation to some in-the-wild scenes with different backgrounds. 
Hence, we propose an event-wise augmentation technique for the background events:  
We capture sequences of both outdoor and indoor scenes without humans with a handheld event camera, \textit{i.e.}, ${\approx}20$ minutes of data, comprising a total of $3.28 \cdot 10^9$ events.
Next, these background scene events are used to 
augment 
the EE3D-S and EE3D-R datasets. 
See Appendix~\ref{subsec:event_aug} 
for details on our event-based augmentation. 

\section{Experimental Evaluation}\label{sec:experiments}  
This section describes our experimental results including numerical comparisons to the most related methods (Sec.~\ref{ssec:comparisons}), an ablation study validating the contributions of the core method modules (Sec.~\ref{ssec:ablative}) as well as comparisons in terms of the runtime and architecture parameters (Sec.~\ref{subsection:runtime}). 
Finally, we show %
a real-time demo 
(Sec.~\ref{ssec:qualitative}).

\noindent \textbf{Implementation Details.} 
We implement our method in PyTorch~\cite{paszke2019pytorch} and use Adam optimizer~\cite{Kingma2015} with a batch size of $27$. 
We first train the network on the EE3D-S dataset with a learning rate of $10^{-3}$ for $8 \cdot 10^5$ iterations and then fine-tune it on the EE3D-R dataset with a learning rate of $10^{-4}$ for $1.5 \cdot 10^4$ iterations. 
All modules of our EE3D architecture are jointly trained. 
The network is supervised using the most recent ground-truth pose within the time window $T$ when constructing the LNES frame, \ie the ground-truth pose is aligned with the latest event in the LNES.
We set $T=15$ms and $N = 20$ for our experiments. 
The performance metrics are reported on a GeForce RTX 3090. 
The real-time demo is performed on a laptop equipped with a 4GB Quadro T1000 GPU, which is housed in a backpack as illustrated in Fig.~\ref{fig:teaser}-(b). 
\par
\noindent \textbf{Evaluation Methodology}. 
We first pre-train EE3D on the EE3D-S dataset and subsequently fine-tune it on EE3D-R. 
The evaluation is conducted on EE3D-R: 
eight subjects are used for pre-training and two subjects each are used for validation and testing. 
Since no existing method addresses 3D human pose estimation from egocentric event streams, we adapt two existing 3D pose estimation methods for our problem setting: 
\begin{itemize} 
\item 
Xu~\etal~\cite{xu2019mo2cap2} and Tome~\etal~\cite{tome2019xr} are egocentric RGB-based methods: We modify their first convolution layer to accept the LNES representation.
\item Rudnev~\etal~\cite{rudnev2021eventhands}, \ie an event-based method that takes LNES as input and estimates hand poses: We modify its output layer to regress 3D human poses.     
\end{itemize}
We follow previous works \cite{xu2019mo2cap2, hakada2022unrealego, zhao2021egoglass} and report the Mean Per Joint Position Error (MPJPE) and MPJPE with Procrustes alignment~\cite{kendall1989survey} (PA-MPJPE).

\subsection{Comparisons to the Related State of the Art}\label{ssec:comparisons} 
\begin{table*}[htbp]
\centering
\resizebox{\textwidth}{!}{
\begin{tabular}{@{}llcccccccccccc@{}}
\toprule
Method & Metric & 
\multicolumn{1}{c}{Walk} &
\multicolumn{1}{c}{Crouch} &
\multicolumn{1}{c}{Pushup} &
\multicolumn{1}{c}{Boxing} &
\multicolumn{1}{c}{Kick} &
\multicolumn{1}{c}{Dance} &
\multicolumn{1}{c}{Inter.~with env.} &
\multicolumn{1}{c}{Crawl} &
\multicolumn{1}{c}{Sports} &
\multicolumn{1}{c}{Jump} &
\multicolumn{1}{c}{Avg. ($\sigma$)} \\
% \cmidrule(l){1-13} 
\hline

{\multirow{2}{*}{Tome \etal \cite{tome2019xr}}} & 

MPJPE &
140.34 &
173.93 &
157.29 &
177.07 &
181.12 &
212.61 &
\multicolumn{1}{c}{169.80} &
144.80 &
207.56 &
165.57 &
173.01 (23.62) \\&

PA-MPJPE & 
104.34 &
119.89 &
102.39 &
124.28 &
121.64 &
132.86 &
\multicolumn{1}{c}{111.89} &
88.94 &
120.15 &
110.32 &
113.67 (12.76) \\

\hdashline

{\multirow{2}{*}{Xu \etal \cite{xu2019mo2cap2}}} & 

MPJPE &
86.09 &
\textbf{153.53} &
199.34 &
133.15 &
114.00 &
104.44 &
\multicolumn{1}{c}{114.52} &
187.95 &
128.21 &
114.10 &
133.53 (36.42) \\&

PA-MPJPE & 
59.11 &
113.31 &
147.13 &
102.50 &
91.75 &
79.65 &
\multicolumn{1}{c}{85.83} &
138.12 &
98.10 &
89.19 &
100.47 (26.52) \\

% \cmidrule(l){1-13} 
\hdashline
\multirow{2}{*} {Rudnev \etal \cite{rudnev2021eventhands}} & 

MPJPE &
74.82 &
178.23 &
105.68 &
\textbf{128.93} &
112.45 &
98.14 &
\multicolumn{1}{c}{110.05} &
120.51 &
110.16 &
106.19 &
114.52 (26.54) \\&

PA-MPJPE & 
56.77 & 
108.34 &
84.15 &
\textbf{100.39} &
91.84 &
78.16 &
\multicolumn{1}{c}{74.62} &
83.47 &
84.83 &
86.09 &
84.87 (14.08) \\

% \cmidrule(l){1-13} 
\hdashline
\multirow{2}{*}{EventEgo3D (Ours)} &

MPJPE &
\textbf{70.88} &
163.84 &
\textbf{97.88} &
136.57 &
\textbf{103.72} &
\textbf{88.87} &
\multicolumn{1}{c}{\textbf{103.19}} &
\textbf{109.71} &
\textbf{101.02} &
\textbf{97.32} &
\textbf{107.30} (25.78) \\&

PA-MPJPE & 

\textbf{52.11} &
\textbf{99.48} &
\textbf{75.53} &
104.66 &
\textbf{86.05} &
\textbf{71.96} &
\multicolumn{1}{c}{\textbf{70.85}} & 
\textbf{77.94} &
\textbf{77.82} &
\textbf{80.17} &
\textbf{79.66} (14.83) \\

\bottomrule
\end{tabular}%
}
\caption{
Numerical comparisons on the EE3D-R dataset (in $mm$). Our EventEgo3D outperforms
existing approaches on most activities by a substantial margin and achieves $6\%$ improvement over Rudnev \etal \cite{rudnev2021eventhands}. ``$\sigma$'' denotes the standard deviation of MPJPE or PA-MPJPE. %\CMh{All metrics are in mm.}
% for various motions. 
}
\label{tab:sotabenchmark_full}
\end{table*}

Table \ref{tab:sotabenchmark_full} presents quantitative results of our approach and compared methods~\cite{xu2019mo2cap2, rudnev2021eventhands, tome2019xr} adapted for our setting.
EE3D outperforms Xu~\etal~\cite{xu2019mo2cap2}, Rudnev~\etal~\cite{rudnev2021eventhands} and Tome~\etal~\cite{tome2019xr} by a large margin, \eg by $6.30\%$, $19.64\%$ and 
$37.98\%$ on MPJPE on average, respectively.

\begin{figure}[t]
\centering
   \includegraphics[width=1\linewidth]{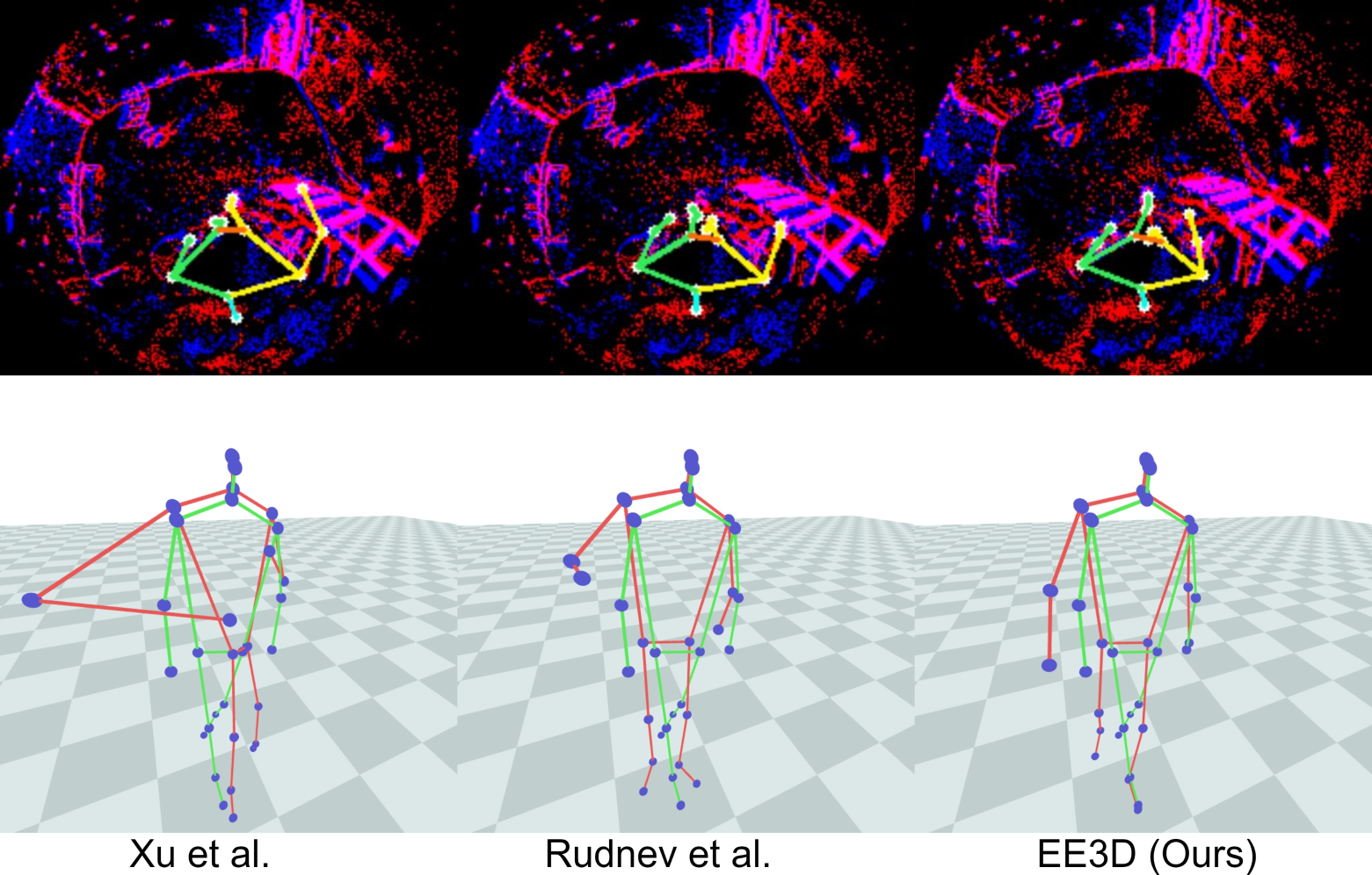}
   \caption{
    Qualitative results of our method in comparison to Xu \etal~\cite{xu2019mo2cap2} and Rudnev \etal~\cite{rudnev2021eventhands}. 
    Note how the previous methods  
    fail to estimate accurate 3D poses
    when events generated by the background become more prevalent than events around the human. 
    The predictions are in red and the ground truth is in green. 
} 
\label{fig:qualitative_compare}
\end{figure}

It is worth noting that our method demonstrates a superiority over the competing methods especially in complex motions involving interaction with the environment, crawling, kicking, sports and dancing.
These motions often come with fast-paced and jittery movements of the HMD, generating substantial background event noise.
Notably, our method excels in handling such challenging scenarios.
Also, we achieve the lowest standard deviation $\sigma$ of the 3D errors on average. 
This result indicates that our method can estimate consistently accurate 
3D poses across various motion activities.
Fig.~\ref{fig:qualitative_compare} shows visual outputs from our approach and compared methods. 
Notably, the events generated by the hand exhibit very close proximity to the events generated by the background. 
Competing methods can not handle such situations,  predicting the background regions as the position of the hand. 
However, our method can accurately estimate 3D poses even in the presence of noisy background events. 
\begin{figure}[t]
\centering
   \includegraphics[width=1\linewidth]{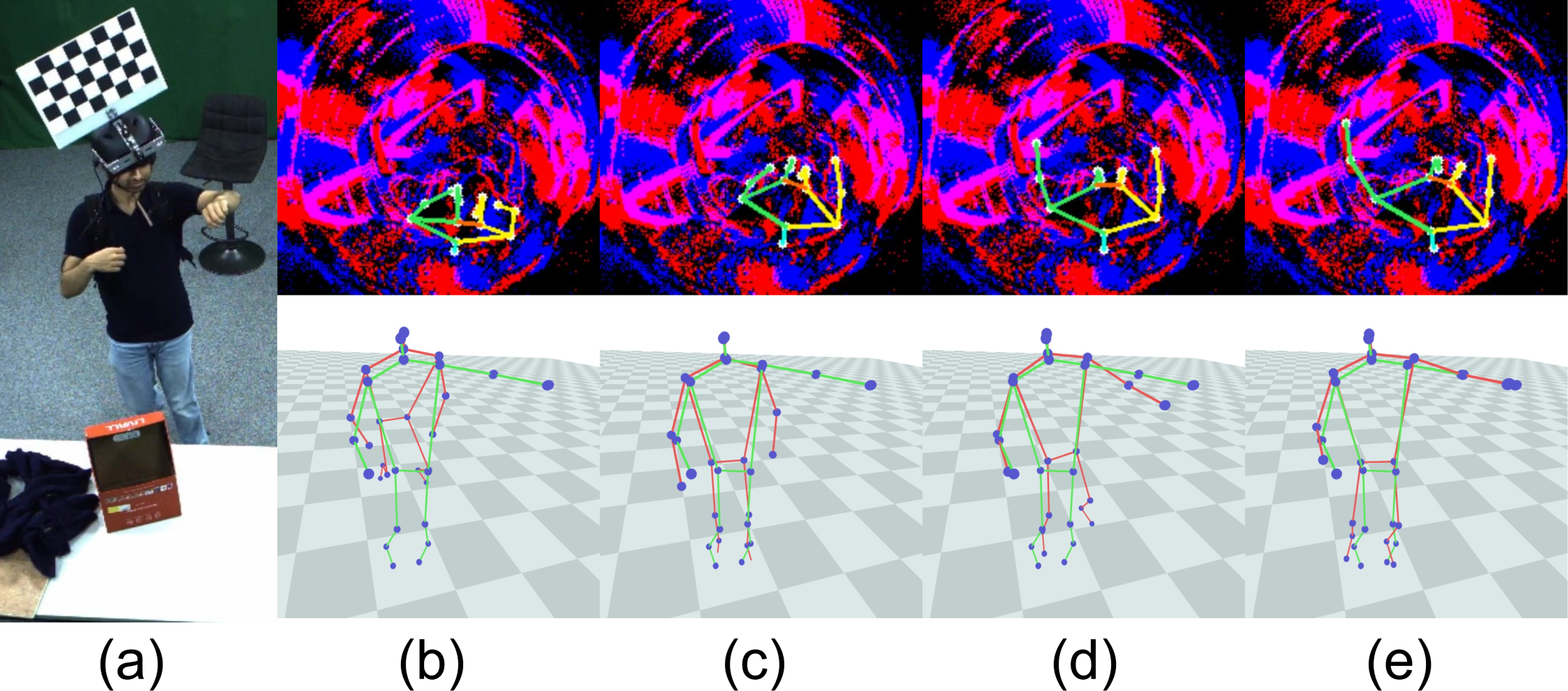}
   \caption{
     Ablation study of REPM on EE3D-R (representative example):  
     (a) Reference RGB view; 
     (b) baseline (EPM only);  
     (c) baseline with segmentation decoder;  
     (d) baseline with REPM without confidence decoder, and 
     (e) EE3D (full model). 
     The predictions are in red and the ground truth is in green. 
} 
\label{fig:ablation}
\end{figure}
\begin{figure*}[t]
\centering
   \includegraphics[width=1\linewidth]{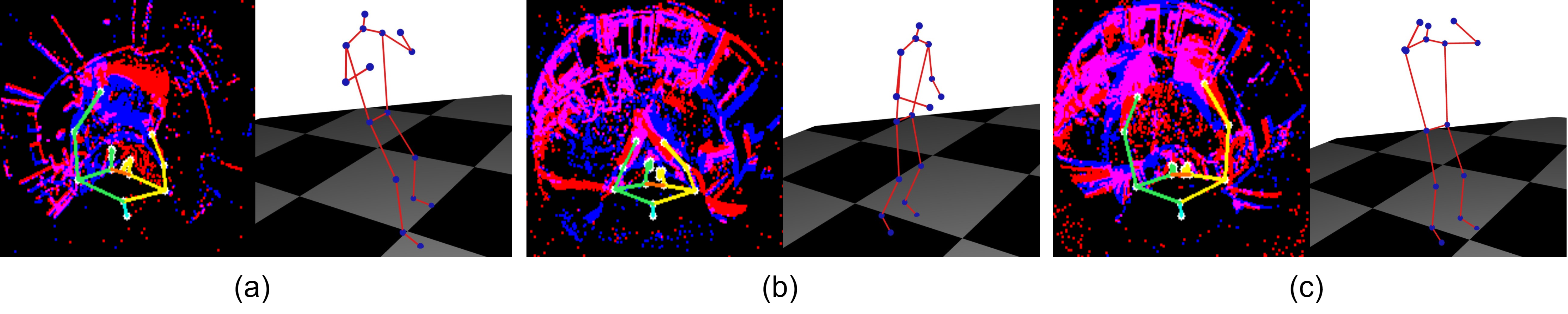}
    \vspace{-8mm}
   \caption{
    Qualitative results of our method on in-the-wild motion sequences: 
         (a) Holding a laptop, 
         (b) Clapping and 
         (c) Waving. 
    The spatiotemporal scene context can be observed in the input LNES (events triggered by the laptop movement in (a) or raised hands in (c)). 
} 
\label{fig:qualitative_results}
\end{figure*}
\subsection{Ablation Study}\label{ssec:ablative} 
We next perform an ablation study to systematically evaluate the contributions of the core modules of our method. 
We define the baseline method as a version with the EPM only. 
We next systematically examine the impact of the REPM as shown in Table \ref{tab:ablation_representation}. 
We see the baseline with the addition of the segmentation decoder improves the performance.
We further notice a performance improvement when we allow past events to propagate to the current frame by weighting the events only with the segmentation decoder.
Finally, including the confidence decoder to weigh the events from the previous frame, yields the best the best MPJPE and PA-MPJPE. 
Fig.~\ref{fig:ablation} shows the 
the effectiveness of the REPM. 
We observe the baseline method's susceptibility to background events Fig.~\ref{fig:ablation}-(a).
Although adding the segmentation decoder aids in mitigating this issue, it still struggles to estimate the correct hand position Fig.~\ref{fig:ablation}-(b). 
Residual events from the previous frame weighted by the human body mask result in a significant performance improvement Fig.~\ref{fig:ablation}-(c).
Finally, our full framework with 
the confidence decoder provides the closest possible estimate for the 3D pose in comparison to the ground truth Fig.~\ref{fig:ablation}-(d).
\begin{table}[htbp] 
\centering
\resizebox{\columnwidth}{!}{
\begin{tabular}{@{}lccc@{}}
\toprule
   & MPJPE $\downarrow$ & PA-MPJPE $\downarrow$ \\
 \hline
 \text{Baseline (EPM only)}                                    &  111.01           &         85.58  \\
 \text{Baseline with segmentation decoder}          &  108.85           &         84.98  \\
 \text{Baseline with REPM w/o Confidence decoder}    &  107.58           &         83.95  \\
 \text{EventEgo3D (Ours)}                                 &  \textbf{107.30}  & \textbf{79.66}  \\ 
\bottomrule
\end{tabular}
}
\caption{%
Ablation study on the EventEgo-R dataset.}
\label{tab:ablation_representation} 
\end{table}

\subsection{Runtime and Performance} \label{subsection:runtime}
\begin{table}[htbp] 
\centering
\resizebox{\columnwidth}{!}{
\begin{tabular}{@{}lccc@{}}
\toprule
   & Params $\downarrow$ & FLOPs $\downarrow$ & Pose Update Rate $\uparrow$ \\
 \hline

\text{Tome \etal~\cite{tome2019xr}} & 
77.01M      &  11.46G  &  77.07
\\
\text{Rudnev \etal~\cite{rudnev2021eventhands}} & 
11.2M & 3.58G & \textbf{489.56}
\\
\text{Xu \etal~\cite{xu2019mo2cap2}} &  
82.18M & 44.06G & 68.65 \\
\text{EventEgo3D (Ours)} & 
\textbf{1.25M} & \textbf{416.84M} & 139.88\\
\bottomrule
\end{tabular}
}
\caption{%
Architecture (number of parameters), performance and runtime (pose update rate) comparisons for the evaluated methods. 
}
\label{tab:runtime} 
\end{table}

EventEgo3D supports real-time 3D human pose update rates of $140$Hz. 
From Table \ref{tab:runtime}, we see that our method has the lowest number of parameters and the lowest number of required floating point operations (FLOPs) compared to the competing methods. 
Rudnev~\etal~\cite{rudnev2021eventhands} is the fastest approach and the second-best in terms of 3D accuracy. 
We achieve the second-highest number of pose updates per second. 
This result highlights that our approach is well-suitable for mobile devices: Its memory and computational requirements as well as power consumption (due to the event camera) would be the lowest among the tested methods. 

Since Rudnev~\etal~\cite{rudnev2021eventhands} use direct regression of 3D joints, their method is faster while our method and Xu~\etal \cite{xu2019mo2cap2} use heatmaps as an intermediate representation to estimate the 3D joints. 
Xu~\etal and Tome~\etal are not designed for event streams
and achieve lower 3D accuracy. 
Moreover, the operations by Rudnev~\etal~are well parallelisable, which explains its high pose update rate. 
\subsection{Real-time Demo}\label{ssec:qualitative} 
Event cameras provide high temporal event resolution and can operate under low-lighting conditions due to their excellent high dynamic range properties. 
Our EE3D approach runs at real-time 3D pose update rates, and we design a real-time demo setup; see Fig.~\ref{fig:teaser}-(b) with a third-person view. 
Our portable HMD enables a wide range of movements, and the on-device computing laptop housed in the backpack allows us to capture in-the-wild sequences. 

We showcase two challenging scenarios, \textit{i.e.}~with fast motions and in a poorly lit environment that would lead to increased exposure time and motion blur in images captured by mainstream RGB cameras. 
Moreover, Fig.~\ref{fig:qualitative_results} illustrates some of the challenging motions performed during the demo, highlighting that our method accurately estimates 3D poses for each motion. 
Notably, in Fig.~\ref{fig:qualitative_results}-(c), a fast-paced waving motion is depicted, and our method successfully recovers the 3D poses in this dynamic scenario.
See our video with the recordings of the real-time demo. 

\vspace{10pt}
\section{Conclusion} 
EventEgo3D addresses the new 
problem, \textit{i.e.}, 3D human motion capture from egocentric event cameras, 
and we introduce all the necessary tools required to address fundamental challenges in designing the method (HMD, the synthetic and real datasets; and neural architecture with components tailored to the problem). 
EventEgo3D runs at real-time 3D pose update rates and is experimentally shown as the most accurate approach among all compared methods: Noteworthy
is that the largest improvements in the 3D reconstruction accuracy are observed under the most challenging 
human motions. 
The EE3D-R dataset used for fine-tuning our model---initially trained on synthetic data---helps to bridge the gap between synthetic and real data. 

We conclude that the usage of event cameras in the egocentric 3D human pose estimation setting is justified and offers many advantages. 
Furthermore, we believe egocentric event-based 3D vision in general has a high potential in related fields that are yet to be explored in follow-up works. 
\noindent\textbf{Acknowledgement.}
This work was supported by the ERC Consolidator Grant 4DReply (770784). 
Hiroyasu Akada is also supported by the Nakajima Foundation.

{
    \small
    \bibliographystyle{ieeenat_fullname}
    \bibliography{main}
}

\clearpage
\appendix
\maketitlesupplementary
\renewcommand{\thetable}{\Roman{table}}  
\renewcommand{\thefigure}{\Roman{figure}} 
\setcounter{figure}{0}
\setcounter{table}{0}

This supplementary document first provides detailed information about our datasets in Section \ref{sec:datasets_details}. 
Section \ref{sec:impl_details} discusses the architecture of EE3D and the real-time implementation of our framework.
Next, we offer in Section \ref{sec:additional_exps} a thorough evaluation of our method alongside competing approaches on the test set of EE3D-S and 
conduct an ablation study to analyse various dataset training strategies. 
We also show visualisations of the predictions generated by our method, along with intermediate representations such as human body masks, confidence maps, and 2D joint heatmaps. 
Please check our video for more visualisations\footnote{\url{https://www.youtube.com/watch?v=jatNH0s_k_E}}.

\section{Dataset Details}
\label{sec:datasets_details}

\subsection{EE3D-S (Synthetic)}
\label{subsec:ee3d-s}
Given the difficulty of capturing a large amount of training data for our egocentric setting, we create a synthetic dataset, EE3D-S, for pre-training our method. 
We generate EE3D-S by rendering sequences of human motions using a virtual replica of our HMD. 
For each motion sequence, we first fit the 3D human model SMPL \cite{SMPL:2015} to the egocentric observations of the HMD wearer. 
Subsequently, we animate the human model by sampling from the CMU~MoCap~dataset~\cite{cmumocap}.
Next, we obtain RGB frames and human body masks by rendering the scene from the viewport of the virtual HMD. 
Additionally, the 3D body joint positions of the wearer are obtained by transforming the SMPL body joint positions in the world coordinate frame to the coordinate frame of the HMD. 
Finally, the rendered RGB frames are passed to VID2E~\cite{Gehrig_2020_CVPR} that converts an RGB frame sequence to an event stream, resulting in the EE3D-S dataset. 
To represent background events within the event stream generation process, we model the background by randomly selecting images from the LSUN~dataset~\cite{yu15lsun}.

\noindent \textbf{Scene Modelling.} 
Following Xu~\etal~\cite{xu2019mo2cap2}, we build our dataset on top of the large-scale synthetic human dataset, SURREAL~\cite{Varol:CVPR:2017} using Blender~\cite{blender_soft}.
SMPL~\cite{SMPL:2015} provides the proxy geometry of the HMD wearer, and body textures are randomly sampled from the texture set provided by the SURREAL dataset.
The background is modelled with a $26m^2$ sized plane with textures randomly sampled from the LSUN dataset~\cite{yu15lsun}. 
We illuminate the scene with four point light sources with random positions within a five-meter radius from the HMD to create variously illuminated scenes. 

\noindent \textbf{Human Animation.} The motions of the 3D human model are sampled from the CMU MoCap~\cite{cmumocap} dataset.
While generating events, it is essential to ensure that VID2E has enough temporal information from the motion sequence, as highlighted in Gehrig~\etal~\cite{Gehrig_2020_CVPR}.
Hence, we sample the motions at high frame rates, i.e.~the SMPL body parameters obtained from SURREAL are linearly interpolated to sample motions at $480$Hz.

\begin{figure}[!t]
\centering
   \includegraphics[width=0.99\linewidth]{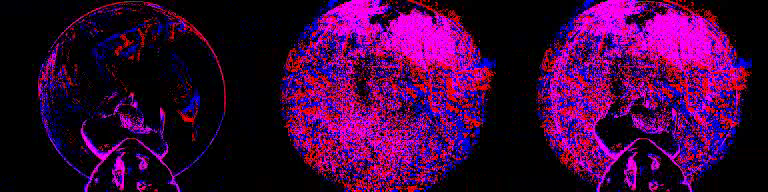}
   \caption{
    Exemplary events from EE3D-S (left), background events (middle) and background events augmented with the events from EE3D-S (right). 
}
\label{fig:dataset_aug}
\end{figure}

\label{ref:hbm_render}
\noindent \textbf{Rendering and Event Stream Generation.} 
We render the scenes using the fisheye camera from the virtual replica of the HMD.
The position of the fisheye camera is obtained by offsetting the nose vertex position of the SMPL body aligning it closely with the event camera mounted on the real HMD, i.e.~the offset is determined by visually aligning the position of the real event camera with respect to the wearer. 
To set the intrinsic parameters of the fisheye camera, we calibrate our real event camera using the omnidirectional camera calibration toolbox OCamCalib~\cite{scaramuzza2006toolbox}.
The obtained intrinsic parameters are then set for the virtual fisheye camera.
Due to the different head sizes and HMD movements during its operation, the camera position with respect to the head can change slightly. 
To account for this effect, we add random perturbations to the position of the virtual fisheye camera. 
We generate the RGB frames and the human body masks using image and mist render layers in Blender's Cycles renderer~\cite{blender_soft}. 
The rendered RGB frames are then processed by VID2E~\cite{Gehrig_2020_CVPR} to generate the event streams.
In total, we synthesise $946$ motion sequences with $6.21 \cdot 10^6$ 3D human poses and $1.419 \cdot 10^{11}$ events.

\begin{figure}[!t]
\begin{center}
   \includegraphics[width=1.0\columnwidth]{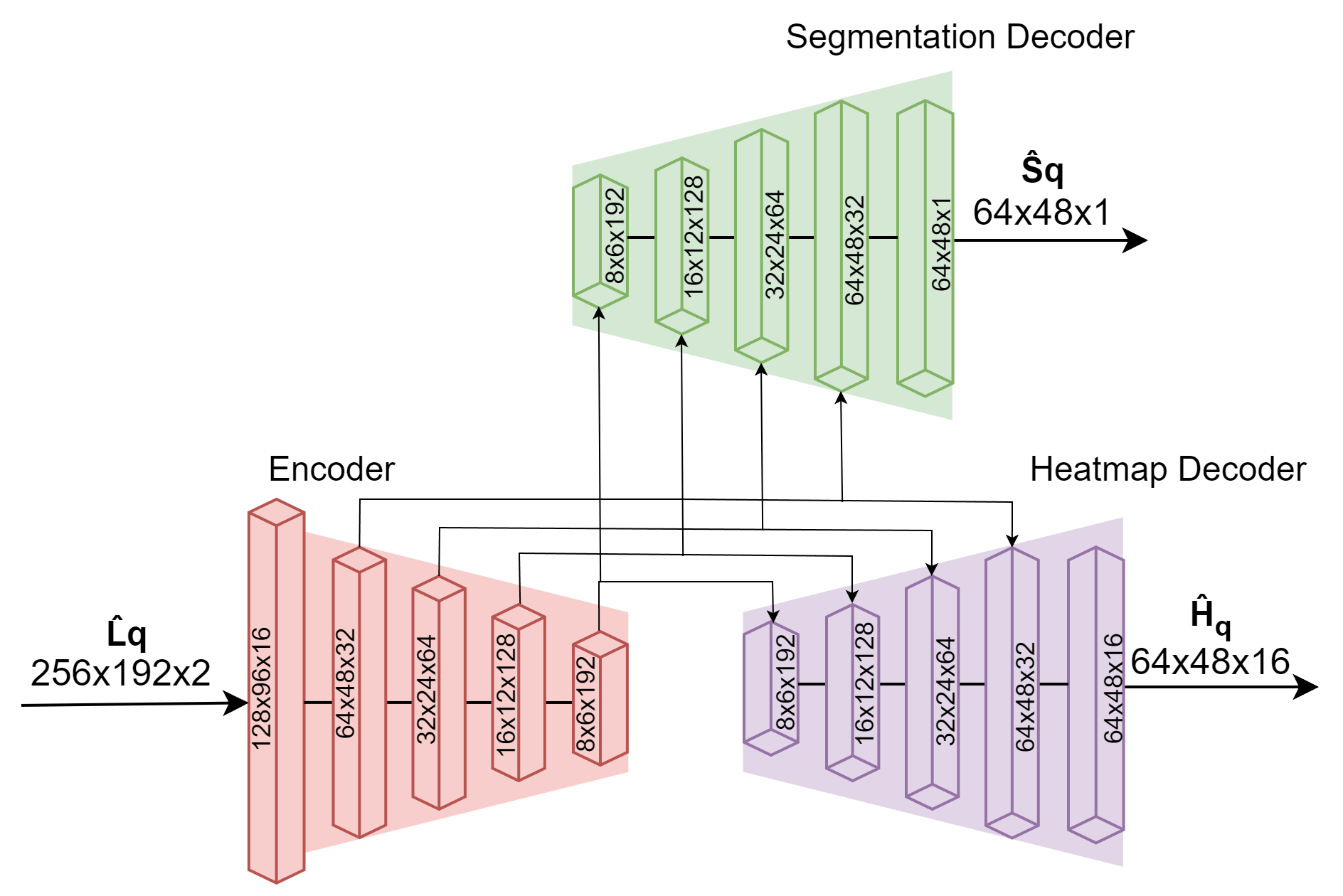}
\end{center}
    \vspace{-10pt} 
    \caption{
        Network architecture of the encoder (bottom-left), heatmap decoder (bottom-right) and segmentation decoder (top). 
    } 
\label{fig:na_u_net}
\end{figure}
\begin{figure}%
\begin{center}
   \includegraphics[width=0.9\columnwidth]{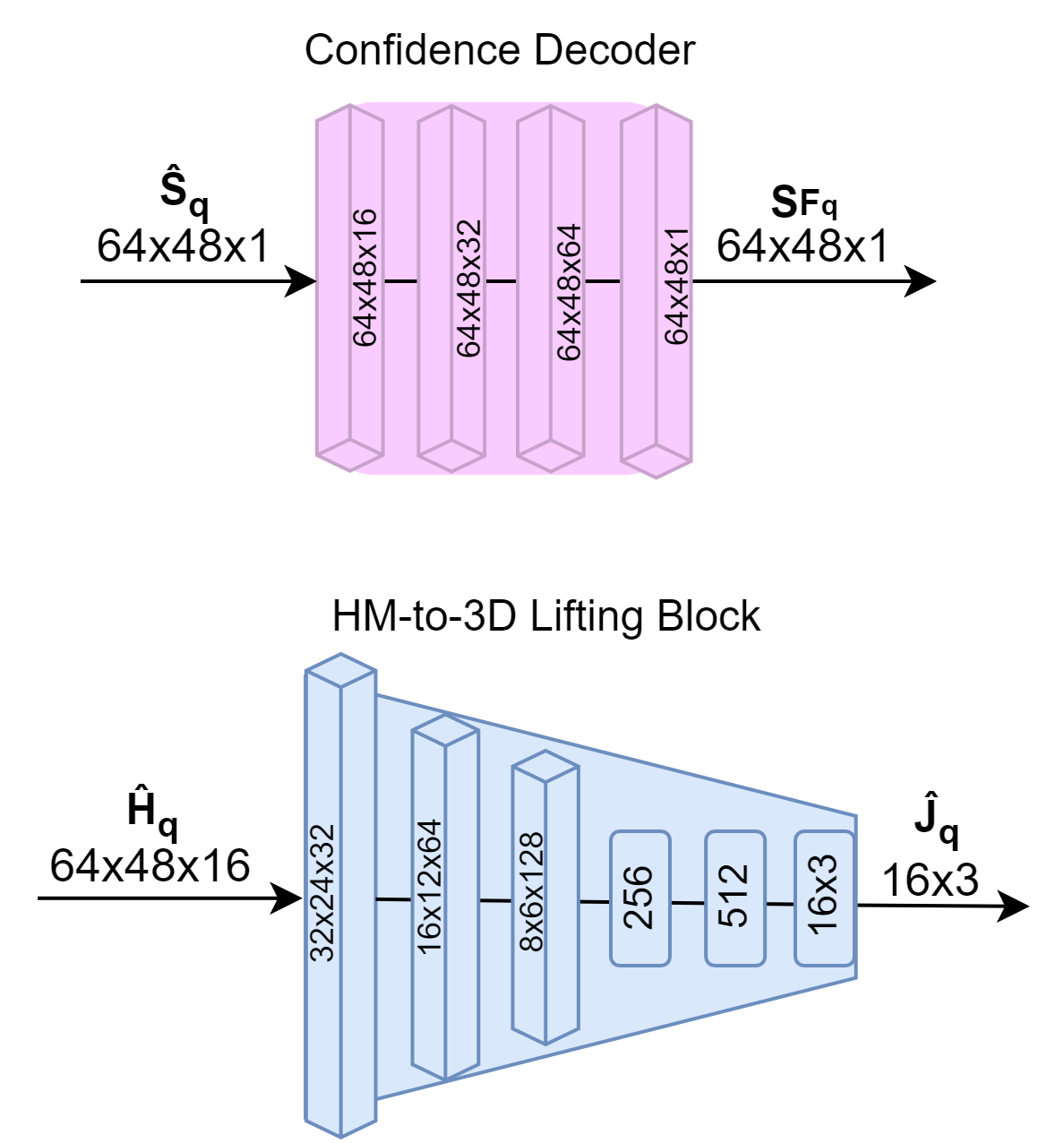}
\end{center}
    \vspace{-10pt} 
    \caption{
        The network architecture of the confidence decoder (top) and Heatmap-to-3D (HM-to-3D) lifting block (bottom).} 
\label{fig:na_cd_hm}
\end{figure}

\subsection{EE3D-R (Real)} 
To evaluate our method and reduce the domain gap between synthetic and real scenarios, we collect the EE3D-R dataset. 

\noindent \textbf{Dataset Composition.} 
The EE3D-R dataset mainly includes regular body movements, featuring a wide range of (natural and unrestricted) motions and inherent differences in their execution among the participants. 
We ask twelve subjects---persons with different body shapes and skin tones---to wear our HMD and perform different motions (e.g.~fast) in a multi-view studio with $29$ RGB cameras recording at $50$ fps. 
Each sequence encompasses the following motion types: Walking, crouching, push-ups, boxing, kicking, dancing, interaction with the environment, crawling, sports and jumping. 
In the sports category, participants perform specific activities---playing basketball, participating in tug of war and playing golf. 
Meanwhile, in the interaction with the environment category, the subjects perform actions such as picking up objects from a table, sitting on a chair, and moving the chair. 
We obtain in total $4.64 \cdot 10^5$ poses spanning ${\approx}155$ minutes.

\noindent \textbf{Human Body Mask Generation.} 
The human body mask is generated in two steps. 
First, we track an SMPL mesh on the wearer of the HMD using open-source software EasyMoCap \cite{easymocap}. 
Subsequently, we transform the tracked SMPL mesh into the local coordinate system of the HMD and then project it onto the egocentric view.

\subsection{Event Augmentation}
\label{subsec:event_aug}
Motions and data recorded in the multi-view studio would not allow satisfactory generalisation to some in-the-wild scenes with different backgrounds. 
Hence, we propose an event-wise augmentation technique for the background events and apply it to EE3D-S and EE3D-R; see Fig.~\ref{fig:dataset_aug}. 

We capture sequences of both outdoor and indoor scenes without humans with a handheld event camera and obtain a background event stream. 
The event-wise augmentation augmentation is done in two steps.
First, we convert the background event stream to LNES frames, each denoted by $\mathbf{L_B}$ with a time window of duration $T$ (refer to Sec. \ref{sec:experiments}). 
We obtain the LNES frame $\mathbf{L_q}$ and its corresponding human segmentation mask from EE3D-S or EE3D-R. 
We next perform element-wise multiplication between $\mathbf{L_B}$ and the inverse of the human segmentation mask, resulting in $\mathbf{L_A}$. 
The inverted human segmentation mask functions as the background mask effectively removing the human.
Finally, we perform element-wise addition of $\mathbf{L_A}$ and   $\mathbf{L_q}$, which serves as the input to our network.

\section{Implementation and Architecture Details}
\label{sec:impl_details}

\subsection{Architecture Details} 

\begin{table*}[!ht]
\centering
\resizebox{\textwidth}{!}{
\begin{tabular}{@{}llcccccccccccc@{}}
\toprule
Method & Metric & 
\multicolumn{1}{c}{Walk} &
\multicolumn{1}{c}{Crouch} &
\multicolumn{1}{c}{Pushup} &
\multicolumn{1}{c}{Boxing} &
\multicolumn{1}{c}{Kick} &
\multicolumn{1}{c}{Dance} &
\multicolumn{1}{c}{Inter. with env.} &
\multicolumn{1}{c}{Crawl} &
\multicolumn{1}{c}{Sports} &
\multicolumn{1}{c}{Jump} &
\multicolumn{1}{c}{Avg. ($\sigma$)} \\
\hline

{\multirow{2}{*}{Ours with EE3D-S}} & 

MPJPE &
326.17 & 
319.17 &
239.92 &
254.62 &
318.03 &
274.43 &
279.79 &
327.40 &
317.49 &
316.91 &
297.39 (32.29) \\&

PA-MPJPE & 
180.83 &
173.56 &
134.47 &
130.56 &
181.62 &
146.33 &
162.28 &
162.90 &
153.61 &
154.39 &
158.05 (17.76) \\

\hdashline

{\multirow{2}{*}{Ours with EE3D-R}} &
MPJPE &
87.21 &
163.80 &
101.40 &
132.50 &
111.68 &
98.72 &
102.07 &
135.71 &
107.53 &
106.27 &
114.69 (22.73) \\&

PA-MPJPE &

69.16 &
110.14 &
78.92 &
103.18 &
94.30 &
77.49 &
72.59 &
104.63 &
81.75 &
82.75 &
87.49 (14.48) \\

\hdashline
\multirow{2}{*} {Ours w/o Augm.} & 

MPJPE &
80.02 &
127.62 &
97.68 &
119.92 &
118.06 &
130.22 &
107.27 &
93.78 &
132.21 &
115.37 &
112.21 (17.25) \\&

PA-MPJPE & 
60.04 &
95.74 &
76.33 &
95.54 &
89.71 &
103.02 &
88.22 &
74.07 &
94.72 &
85.77 &
86.32 (12.82) \\

\hdashline
\multirow{2}{*}{Ours} & 

MPJPE &
\textbf{70.88} &
163.84 &
\textbf{97.88} &
136.57 &
\textbf{103.72} &
\textbf{88.87} &
\multicolumn{1}{c}{\textbf{103.19}} &
\textbf{109.71} &
\textbf{101.02} &
\textbf{97.32} &
\textbf{107.30} (25.78) \\&

PA-MPJPE & 

\textbf{52.11} &
\textbf{99.48} &
\textbf{75.53} &
104.66 &
\textbf{86.05} &
\textbf{71.96} &
\multicolumn{1}{c}{\textbf{70.85}} & 
\textbf{77.94} &
\textbf{77.82} &
\textbf{80.17} &
\textbf{79.66} (14.83) \\
\bottomrule
\end{tabular}%
}
\caption{
Numerical comparisons on the EE3D-R dataset with different dataset training strategies. Fine-tuning our method on EE3D-R after pre-training on EE3D-S yields an improvement of $62.26\%$ in the MPJPE.  
}
\label{tab:dataset_ablation}
\end{table*}

The encoder, heatmap decoder and segmentation decoder are constructed using Blaze blocks \cite{bazarevsky2020blazepose}; see Fig.~\ref{fig:na_u_net}. 
The confidence decoder and Heatmap-to-3D (HM-to-3D) lifting block are built using standard convolution filters.

The confidence decoder is a four-layer fully convolutional neural network. 
In each convolution layer, filters with a kernel size of three are utilised, followed by the PReLU activations \cite{He2015}. 
We apply appropriate padding to maintain the spatial dimensions at each network layer so that the network output has the same dimensions as its input.

The HM-to-3D lifting block shown in Fig.~\ref{fig:na_cd_hm} is a six-layer network with three convolution layers and three dense layers. 
Each convolution layer consists of filters with a kernel size of four, followed by batch normalisation and RELU activation function. 
Subsequently, we perform average pooling and flatten the features outputted by the convolution layers. 
Finally, these features are passed to the dense layers to estimate the positions of the joints denoted as $\hat J$.

\subsection{Real-time Inference}

\noindent \textbf{3D Viewer.}
Our method runs locally on the laptop housed in the backpack.
We visualise the results using a separate device running a 3D viewer \cite{easymocap}.
The predictions generated by our method are transmitted to the 3D viewer through network sockets. 
Note that the data transmission 
causes a slight lag in the visualisations, especially during complex and fast motions for which the pose update frequency is very high. 

\noindent \textbf{Temporal Stability.} 
We reduce the jitter generated by our method for real-time inference with the 1€ filter~\cite{casiez20121}. 
For a fair comparison, the same procedure is also performed for all the methods we evaluate.

\section{Additional Experiments}
\label{sec:additional_exps}

\begin{table}[tbp] 
\centering
\resizebox{\columnwidth}{!}{
\begin{tabular}{@{}lccc@{}}
\toprule
   & MPJPE $\downarrow$ & PA-MPJPE $\downarrow$ \\
 \hline
 \text{Tome \etal~\cite{tome2019xr}}                      &  172.14
    &         124.62  \\
 \text{Rudnev \etal~\cite{rudnev2021eventhands}}          &  217.05      &         136.05  \\
 \text{Xu \etal~\cite{xu2019mo2cap2}}                     &  196.39     &         99.07  \\
 \text{EventEgo3D (Ours)}                                 &  \textbf{124.85}  & \textbf{92.58}  \\ 
\bottomrule
\end{tabular}
}
\caption{%
Quantitative evaluation on EE3D-S. 
}
\label{tab:synthetic_test} 
\end{table}

\noindent \textbf{Ablation Study for the Dataset Training Strategy.} 
Table \ref{tab:dataset_ablation} summarises the quantitative evaluation of our method on the EE3D-R dataset using different training strategies. 
Training our method solely on EE3D-S without fine-tuning on EE3D-R yields the poorest performance.
Pre-training our method on EE3D-S and subsequently fine-tuning it on EE3D-R results in a lower MJPJE (denoted as ``Ours w/o Augm.'') compared to training our method exclusively on EE3D-R (denoted as ``Ours with EE3D-R''). 
Specifically, this approach improves the MJPJE by $2.16\%$. 
Furthermore, augmenting the events with background data (refer to Sec.~\ref{subsec:event_aug}) in conjunction with fine-tuning leads to the best MJPJE of $107.30$mm. 

\noindent \textbf{Evaluation on EE3D-S.} Table \ref{tab:synthetic_test} quantitatively evaluates our approach and competing methods on EE3D-S.
In this experiment, all the methods are pre-trained on EE3D-S and fine-tuned on EE3D-R. 
We then compare the methods on the test set of EE3D-S. 
Overall, our method achieves the lowest MPJPE  outperforming Tome~\etal\cite{tome2019xr} and Xu~\etal~\cite{xu2019mo2cap2} by $27.47\%$ and $36.42\%$ on MPJPE, respectively. 
Rudnev~\textit{et al.}'s approach~\cite{rudnev2021eventhands} performs the worst in our testing achieving an MPJPE of $217.05$mm.

\noindent \textbf{Additional Visualisations.} 
We provide additional visualisations for our method showcasing intermediate representations produced by each module in Table~\ref{fig:additionalresults}. 
The input $\mathbf{\hat{L}}_{q}$ is computed based on 
$\mathbf{L}_q$ and $\mathbf{\hat{L}}_{q-1}$, i.e.~the current and previous input LNES frames. 
The segmentation decoder first estimates the segmentation mask that subsequently serves as input for the confidence decoder to generate the confidence map. 
Simultaneously, the heatmap decoder estimates the heatmaps of the human body joints. 
These heatmaps are then passed to the HM-to-3D lifting block resulting in the regression of the 3D joint locations.

\newcolumntype{C}{>{\centering\arraybackslash}m{8.7em}}
\begin{table*}\sffamily
\centering
\begin{tabular}{C*8{C}@{}}
\toprule
Input frame $\mathbf{\hat{L}}_{q}$ & Human Body Mask & Confidence map & Heatmap & Prediction \\

\midrule

\includegraphics[width=8em, trim=0 0cm 0 0]{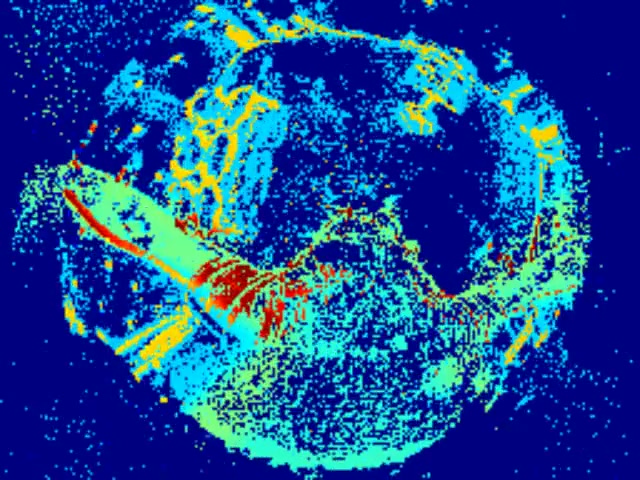}  &
\includegraphics[width=8em]{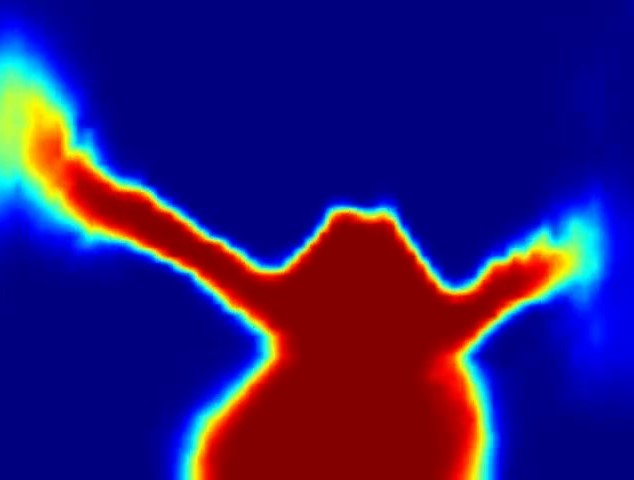} &
\includegraphics[width=8em]{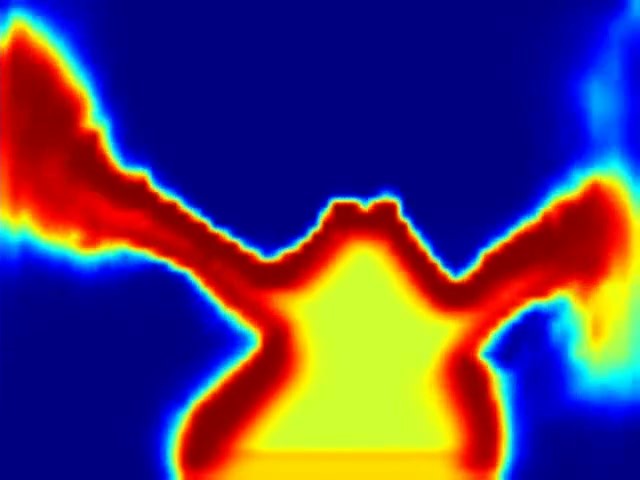} &
\includegraphics[width=8em]{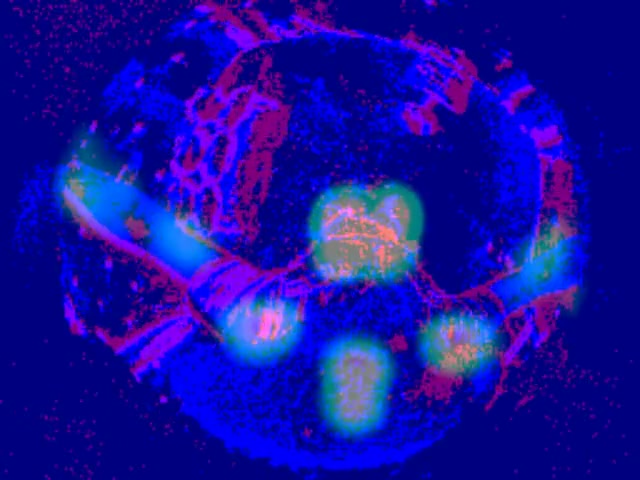} &
\includegraphics[width=8em]{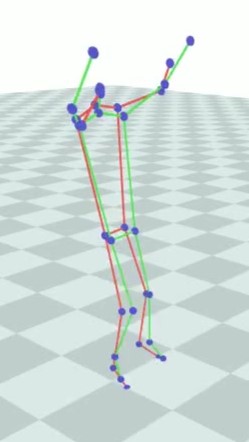} \\

\includegraphics[width=8em, trim=0 0cm 0 0]{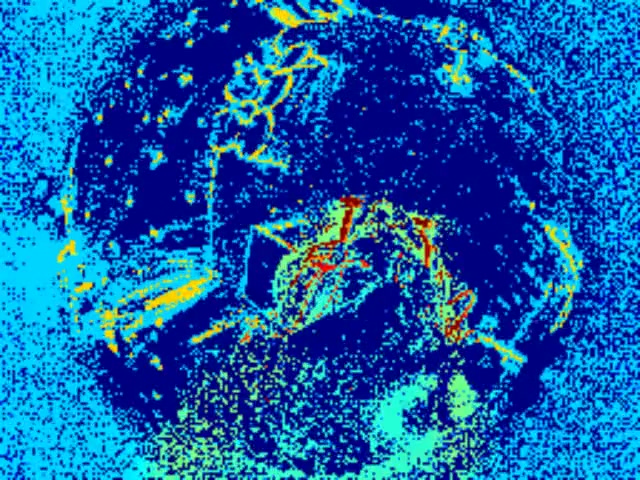}  &
\includegraphics[width=8em]{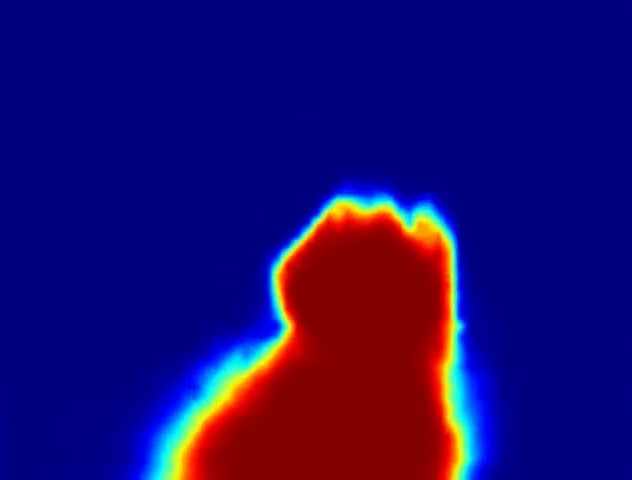} &
\includegraphics[width=8em]{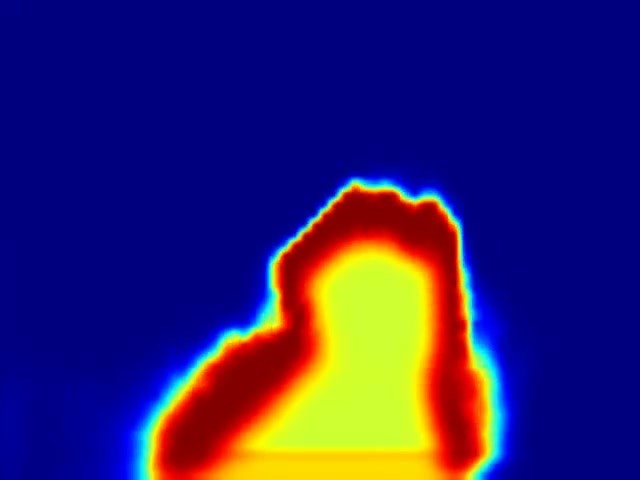} &
\includegraphics[width=8em]{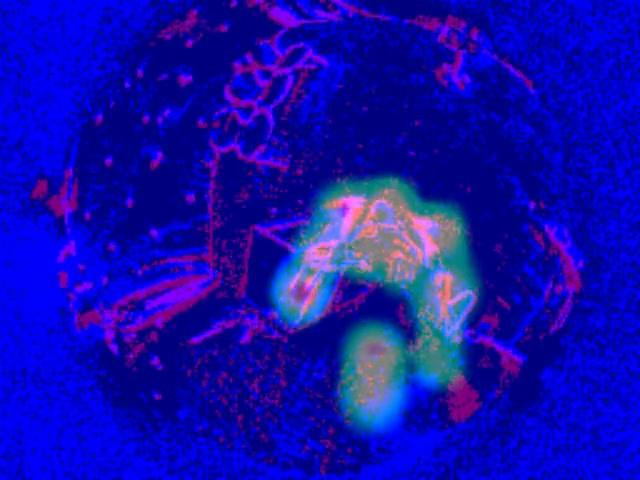} &
\includegraphics[width=8em]{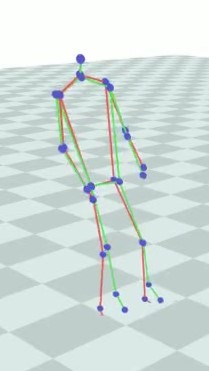} \\

\includegraphics[width=8em, trim=0 0cm 0 0]{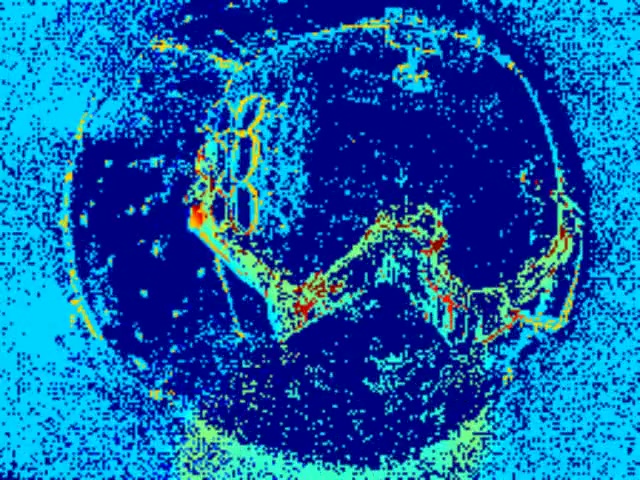}  &
\includegraphics[width=8em]{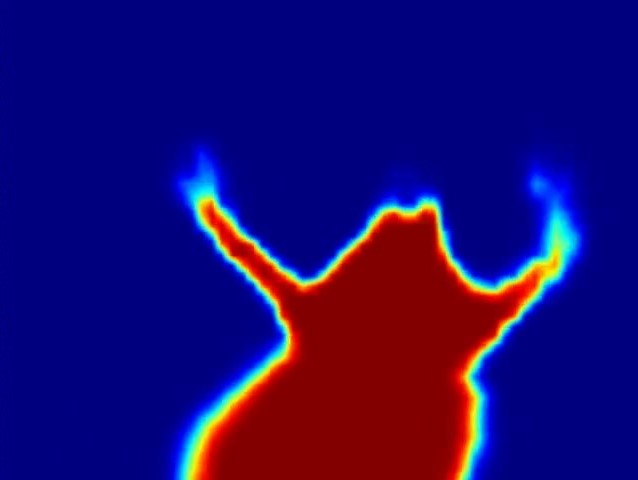} &
\includegraphics[width=8em]{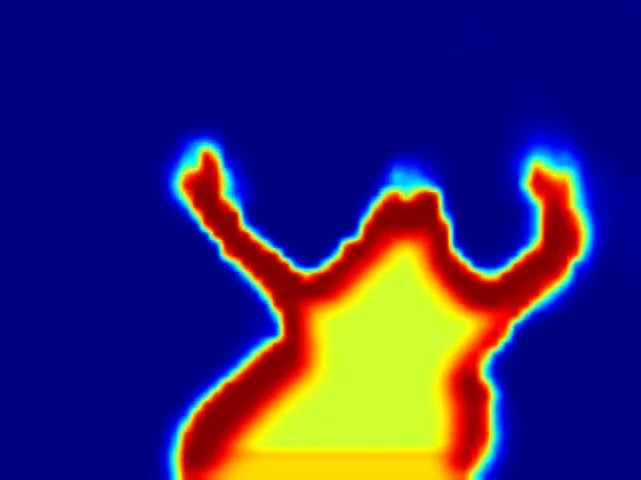} &
\includegraphics[width=8em]{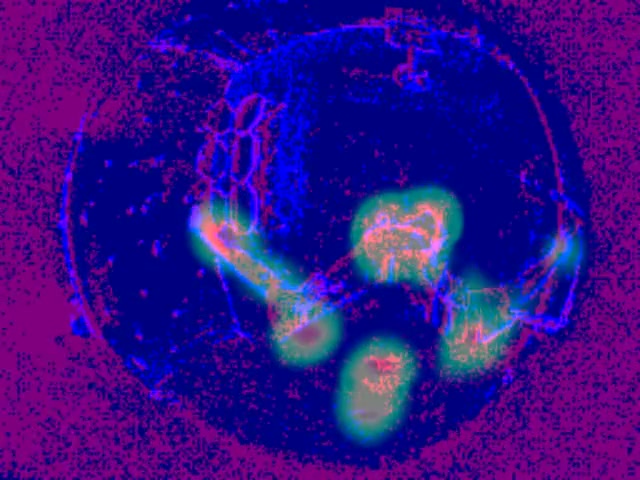} &
\includegraphics[width=8em]{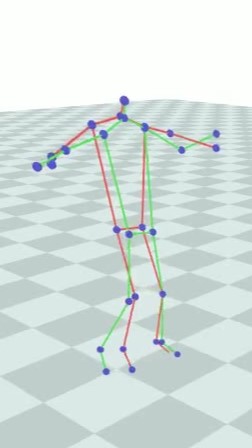} \\

\includegraphics[width=8em, trim=0 0cm 0 0]{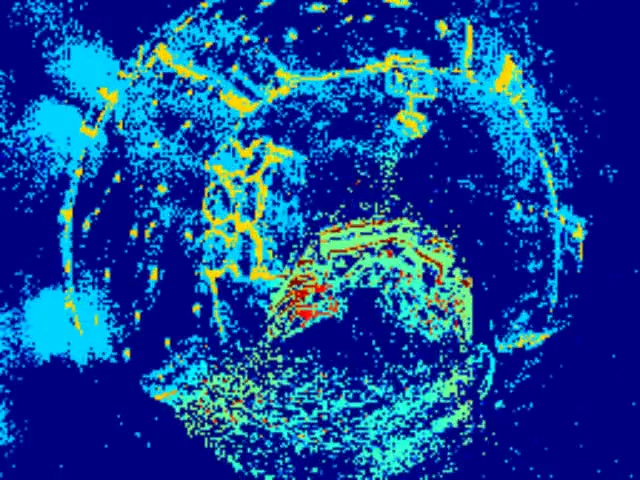}  &
\includegraphics[width=8em]{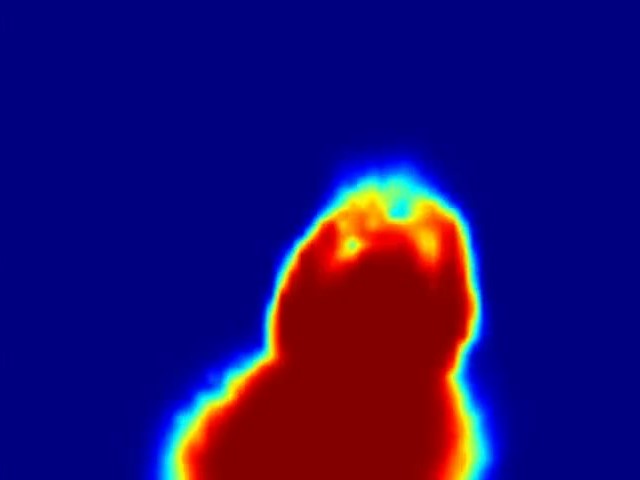} &
\includegraphics[width=8em]{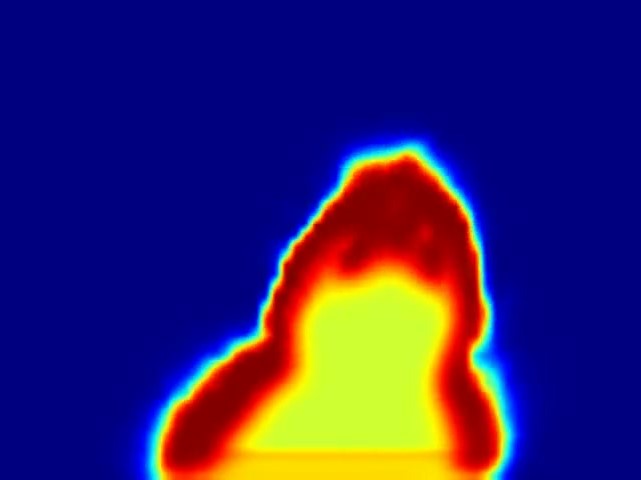} &
\includegraphics[width=8em]{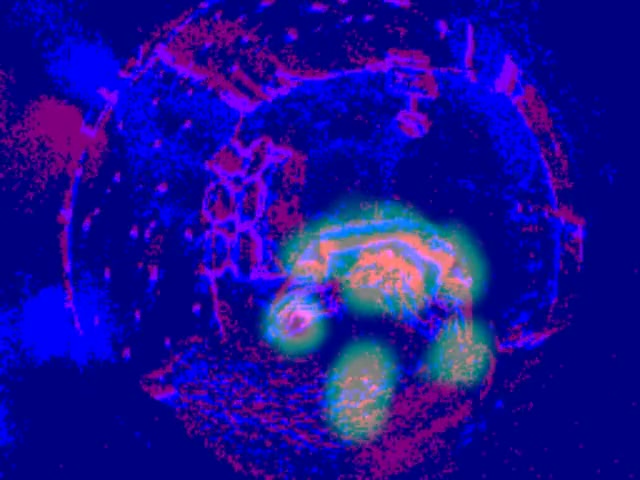} &
\includegraphics[width=8em]{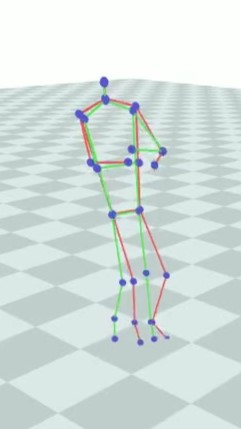} \\

\bottomrule
\end{tabular}
\caption{
Additional visualisations of EE3D along with the visualisations of input frames $\mathbf{\hat{L}}_{q}$, human body masks, confidence maps, heatmaps and the 3D predictions. 
The 3D poses depicted in green represent the ground truth, while those in red signify the predictions by our method. 
The colour scheme for the input frames $\mathbf{\hat{L}}_{q}$, human body masks and confidence maps can be found in Fig.~\ref{fig:ed_l_inp} of the main paper. 
}
\label{fig:additionalresults}
\end{table*}

\end{document}